\documentclass{article}

\PassOptionsToPackage{numbers, compress}{natbib}
 \usepackage[preprint]{neurips_2026}

\usepackage[utf8]{inputenc} 
\usepackage[T1]{fontenc}    
\usepackage{hyperref}       
\usepackage{url}            
\usepackage{booktabs}       
\usepackage{amsfonts}       
\usepackage{nicefrac}       
\usepackage{microtype}      
\usepackage{xcolor}         

\usepackage{config}
\usepackage{subfiles}

\title{Deep Psychovisual Image Representations}

\author{%
 Wendi Ma \\
 School of EECS \\
 The University of Queensland \\
 \texttt{wendi.ma@uq.edu.au} \\
 \And
 Aryaman Sharma \\
 School of EECS \\
 The University of Queensland \\
 \texttt{aryaman.sharma@uq.edu.au} \\
 \And
 Wei Dai\\
 School of EECS \\
 The University of Queensland \\
 \texttt{wei.da@uq.edu.au} \\
 \And
 Shekhar S. Chandra\\
 School of EECS \\
 The University of Queensland \\
 \texttt{shekhar.chandra@uq.edu.au} \\
 }

\begin{document}

\maketitle

\begin{abstract}
    Psychovisual models suggest human vision decouples low-level feature extraction from higher cognition by first forming intermediate abstractions. In contrast, deep learning-based vision models routinely extract and aggregate features using homogeneous stacks of spatial layers, rendering their decision-making processes opaque. In this paper, we propose \acf{DVC}, a learned frequency-domain representation inspired by 1990s image codes that quantised perceptually salient frequencies, which together with complex-valued image representations produces psychovisual-style abstractions. This approach enables the first psychovisual-based deep learning framework, utilizing data-driven spectral filters that learn to encode task-relevant semantic structures within distinct frequency sub-bands. Salience analyses reveal that our psychovisual models extract highly interpretable object parts compared to the amorphous regions produced by regular \acp{CNN}. Furthermore, we find that our models are less depth dependent than \acp{CNN} for model scaling, since our complex-valued representations and learned abstractions subsume the role of deep spatial layers. Together, these findings demonstrate that psychovisual coding provides a promising path toward more efficient and transparent vision models.
\end{abstract}

\begin{figure}[ht]
    \centering
    \includegraphics[width=0.98\linewidth]{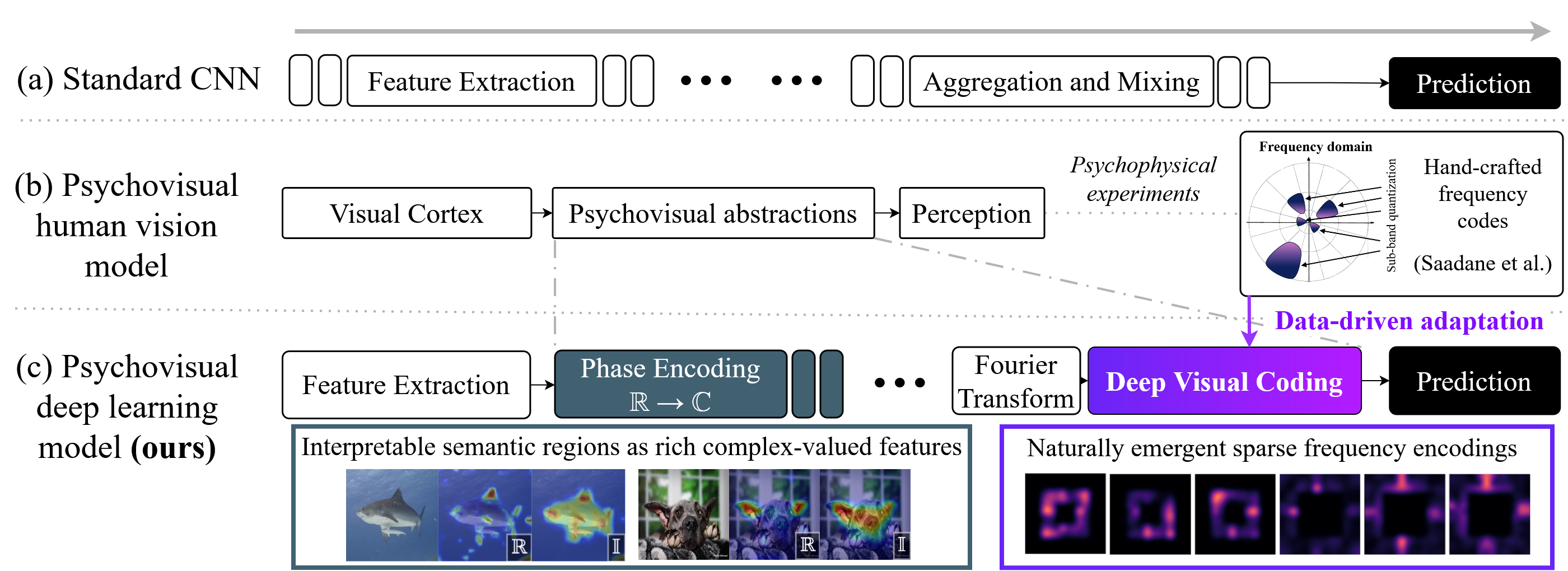}
    \caption{(a) Standard \acp{CNN} use deep stacks of homogeneous spatial layers. (b) \textbf{Psychovisual} models suggest human vision uses explicit intermediate abstractions. (c) Our psychovisual deep learning framework first produces rich complex-valued representations of interpretable semantic regions. They are then encoded in the frequency domain by \textbf{\acf{DVC}}, a data-driven adaptation of hand-crafted psychovisual coding schemes, to introduce psychovisual-like abstractions into deep learning models.}
    \label{fig:intro}
\end{figure}


\newpage
\acresetall
\section{Introduction}

Ever since the ImageNet challenge popularised deep learning for computer vision \citep{alexnet,imagenet}, architectural advances have focused on the design of spatial domain feature extractors, from convolution layers \citep{resnet,resnext,densenet,convnext} to more recent token mixers based on attention mechanisms \citep{vit,mlpmixer,can,gfnet}. These models refine and aggregate image features through deep, homogeneous stacks of spatial layers, where feature processing emerges implicitly through training (Figure \ref{fig:intro} (a)). As a result, intermediate representations are typically produced ad hoc with little explicit structure or organisation. Psychovisual processing (Figure \ref{fig:intro} (b)), the way human vision encodes and interprets visual information, separates feature extraction from higher cognition using intermediate abstractions, such as objects, relations, and categories, providing a natural basis for reasoning \citep{brain_representations_1, brain_representations_2, concept_cells, monkeysee}. 
In the 1990s, psychovisual coding schemes were developed that targeted the perceptually salient frequencies found by human vision studies \citep{saadane_design_1994,guedon_psychovisual_1995,saadane_visual_1998} (see Appendix~\ref{app:psychovisual} for a brief introduction). We hypothesize that these perceptually salient frequency selections may implicitly capture the high-level semantic abstractions utilized by human cognition.

In this work, we develop a data-driven adaptation of psychovisual coding that introduces abstraction-like representations into deep learning frameworks (Figure \ref{fig:intro} (c)). To the best of our knowledge, our work represents the first data-driven exploration of the frequency domain for high-level representation learning in vision, whereas previous studies primarily focused on lower-level feature learning or parameterizing spatial models \citep{chi_fast_2020, rippel_spectral_2015, fccn}. Furthermore, our approach processes feature abstractions almost entirely in the frequency domain without the need for repeated spatial-to-spectral transformations. To summarise, the key contributions of this work are:

\begin{itemize}
    \item We propose \textbf{\acf{DVC}}, a first-of-its-kind data-driven adaptation of psychovisual coding schemes designed to introduce human-vision-like abstractions into deep learning models. It employs learnable, band-limited frequency filters that learn representations of high-level semantic image information, emerging as sparse selections of coronal frequency sub-bands in the \ac{DFT} (see Appendix~\ref{app:background_freq} for a brief introduction).
    \item We introduce \PhasorBlock which has the ability to learn complex-valued features as phase information from real-valued coloured signals to build complex-valued image features that are well suited to the Fourier domain and the proposed \ac{DVC}.
    \item We assemble \ac{DVC} and Phasor Blocks into a cohesive psychovisual deep learning framework. We demonstrate that this pipeline enables interpretable psychovisual-like processing, maintaining or improving performance across multiple classification benchmarks.

    \item We provide salience analysis that demonstrate evidence of semantic decision making that reveal intermediate spatial layers consistently focus on meaningful object parts. These features are then encoded by \ac{DVC} and used for final output prediction, mirroring the use of abstractions to support reasoning in a psychovisual model of human vision.
\end{itemize}

\section{Background}\label{sec:background}

\paragraph{Biologically-inspired Vision.}
Biologically inspired approaches in computer vision predominantly focus on modelling early vision stages. In particular, much attention has been given to \acp{RF} - regions of visual stimuli that elicit strong neural responses in the visual cortex. Mammalian \acp{RF} are known to act as directional differential operators, closely resembling traditional image processing functions like wavelets and Gabor filters \citep{rf_wavelet, rf_cat, rf_macaque}. These parallels motivated their use in approximating low-level human vision, serving as effective feature extractors for basic visual structures like edges and shapes. In deep learning, these functions have been used to build neural networks that mimic cortical pathways \citep{bio_contourlet}, and early-layer \ac{CNN} kernels also perform similar directional operations \citep{alexnet,rippel_spectral_2015}. However, networks that directly incorporate these functions, for example as fixed filter banks or bases for convolution kernels, have generally been restricted to small-scale problems and do not scale well to modern vision tasks \citep{liu_c-cnn_2021}. Beyond \acp{RF}, human vision has also been studied from a frequency domain perspective in previous image coding literature. In particular, our work draws upon psychovisual coding, introduced further below, as the basis for modelling abstractions used in higher vision stages and supports models which do scale to modern tasks.
\paragraph{Frequency Domain Learning.}
Deep learning computer vision models have largely focused on processing in the spatial domain, which expresses localized relationships through contiguous pixel neighborhoods. The frequency (Fourier) domain, conversely, projects signals onto orthogonal basis functions, inherently capturing global representations (a review of the frequency domain and Fourier Transforms is presented in Appendix \ref{app:background_freq}). Formulated in this space, traditional image processing functions like ridgelets~\citep{candes_ridgelets:_1999}, curvelets~\citep{starck_curvelet_2002} and contourlets~\citep{do_contourlet_2005} have appealing sparse representations. In fact, they bear strong resemblances to psychovisual codes since they target specific selections of sub-bands, corresponding to features from different spatial scales.

In deep learning, the frequency domain has primarily been used to exploit the Convolution Theorem \citep{imageprocessingbook}, whereby spatial circular convolution becomes simple element-wise multiplication in the frequency domain. Many works leveraging this property are performance-driven: \citep{falcon,chi_fast_2020,specnet} use frequency domain filters to accelerate \acp{CNN} and incorporate global context, while \citep{gfnet,fnet,huang_aff_2023} employ global frequency filters as lightweight and effective token mixers for transformer-style models. Other studies use frequency domain re-parameterizations of \acp{CNN} to analyse model properties and behaviours \citep{rippel_spectral_2015,niff,kabri_fno_2023}, such as optimal kernel structures. However, while there is extensive work exploiting the frequency domain for computational efficiency and model analysis, it has remained largely unexplored as an explicit representation space in its own right, particularly for high-level semantic features.

\paragraph{Psychovisual Coding.}
Research conducted in the 1990s by Dominique Barba and colleagues used psychophysical experiments to characterise the frequency sensitivities of the human visual cortex \citep{saadane_design_1994,senane_image_1995,saadane_visual_1998}. These findings informed \textit{psychovisual coding}, an image quantization and compression scheme designed to be perceptually lossless. This approach exploits the human visual system’s varying sensitivity to spatial frequencies by decomposing the frequency domain into a series of coronal sub-bands (Figure \ref{fig:semantic_visual_coding} (left)), each assigned quantization thresholds based on measured sensitivities. This process enables frequencies corresponding to perceptually salient image features to be preserved in full, while others are removed or heavily compressed without affecting perceived image quality. An extended review of psychovisual coding literature is presented in Appendix \ref{app:psychovisual}. 

While psychovisual coding was primarily developed for image compression, we speculate that its quantization process may implicitly capture the underlying abstractions used by human vision. Specifically, by isolating those frequency sub-bands whose information is used to construct the high-level abstraction used by the brain for image understanding, the resulting spectral selections may effectively encode the abstractions themselves. Thus, without directly modelling any neural mechanisms, psychovisual coding could serve as a functional proxy for introducing psychovisual-like abstraction into vision models. To the best of our knowledge, this connection has not yet been explored in deep learning, unlike \acp{RF}.

\section{Method}\label{sec:method}
\begin{figure}[h]
    \centering
    \includegraphics[width=0.9\linewidth]{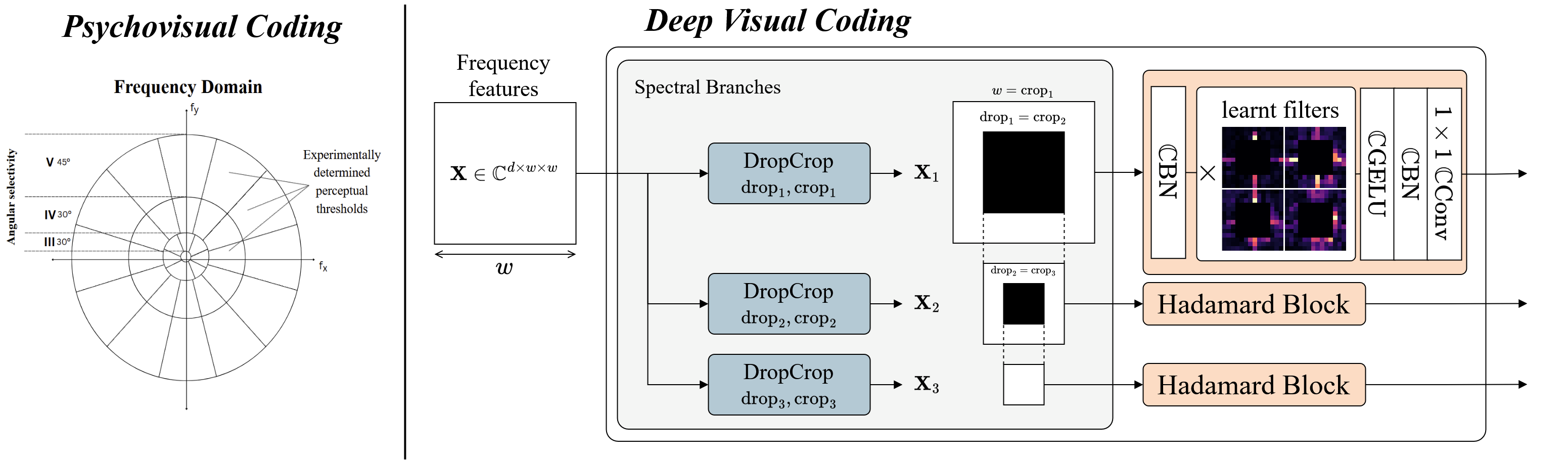}
    \caption{\textbf{(Left) Hand-crafted psychovisual coding} from \citet{saadane_visual_1998}, which quantizes perceptually salient radial frequencies determined by human vision experiments. \textbf{(Right) Our \ac{DVC} module,} a data-driven adaptation of psychovisual coding. It uses (1) \textit{Spectral Branches} for radial spectral decomposition (2) \textit{Hadamard Blocks} to apply learnt element-wise filters and channel mixing. $\mathbb{C}$Conv/BN/GELU denote complex-valued convolution, batch norm and GELU operations - see Appendix \ref{app:psychonet}}
    \label{fig:semantic_visual_coding}
\end{figure}

\paragraph{\acf{DVC}}, shown in Figure \ref{fig:semantic_visual_coding} (right), is a learnable frequency filtering module modelled after the coronal sub-band design of psychovisual coding, intended to realise the psychovisual abstraction step at the core of our approach.



Let $\bm{x} \in \mathbb{C}^{ d \times w \times w}$ be a spatial feature map produced by a neural network, where $d$ is the number of channels and $w\times w$ the spatial size. Its frequency spectrum $\bm{X} \in \mathbb{C}^{ d \times w \times w}$ can be obtained via applying the 2D \ac{DFT} \citep{cooley_finite_1969} on the last two dimensions - a review of the frequency domain and Fourier Transforms is provided in Appendix \ref{app:background_freq}. The \textit{Spectral Branches} module replicates the radial frequency partitioning in psychovisual codes, decomposing $X$ into disjoint rectangular sub-bands $\bm{X}_1, \bm{X}_2, ...$. Each sub-band is produced via applying \textit{DropCrop} blocks, which set a lower frequency boundary ($\text{drop}_i$) by zeroing central frequencies and an upper boundary ($\text{crop}_i$) by cropping $\bm{X}$ to size $d \times \text{crop}_i \times \text{crop}_i$. For each sub-band $X_i$, we then apply a set of learnable filters $\bm{W_i} \in \mathbb{C}^{d \times \textrm{crop}_i \times \textrm{crop}_i}$ via:
\begin{equation}
    {X_i}_{\text{filtered}} = \textrm{Softmax}\br{{X_i} \odot \bm{W_i}}
\end{equation}
where $\odot$ denotes the Hadamard (element-wise) product. Channel-wise softmax is applied here to amplify important frequency selections and suppress unimportant ones - emulating the quantization step of psychovisual coding. This filtering step is encapsulated in a \textit{Hadamard Block} which additionally applies normalization and channel-mixing via a $1\times1$ convolution layer. 

\begin{figure}[ht]
    \centering
    \includegraphics[width=0.9\linewidth]{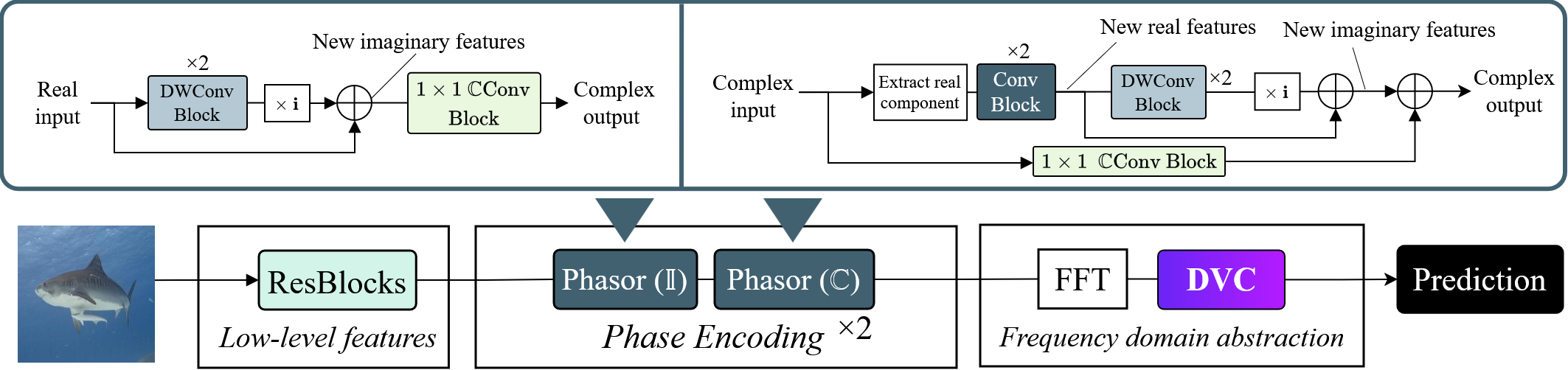}
    \caption{Overall design of the PsychoNet framework (bottom) and Phasor Block architectures (top). ResNet-style Residual Blocks (ResBlocks) first extract low-level features, then are converted into complex-valued representations by Phasor Blocks and encoded in the frequency domain by \ac{DVC}.}
    \label{fig:architecture}
\end{figure}

\paragraph{Phasor Blocks.}
To prepare spatial features for \ac{DVC}, we also introduce Phasor Blocks (Figure \ref{fig:architecture} (top)) to first convert real-valued image features into rich complex-valued representations. As demonstrated in our experiments (Section \ref{sec:representation_analysis}), this transformation produces activations that target interpretable semantic regions, such as characteristic object parts, which provide the part-level features that \ac{DVC} subsequently encodes into psychovisual abstractions. An additional practical motivation for this design is to avoid conjugate symmetry. Real-valued features, such as natural images, incur the conjugate symmetry of the \ac{FT}, which renders half of the frequency domain redundant and limit learnt filtering from fully exploiting complex representations. By generating complementary imaginary components, Phasor Blocks break conjugate symmetry and improve the specificity of learned sub-bands.

In practice, imaginary components are generated from existing real features using lightweight depthwise convolution based blocks which decouple spatial and channel mixing to encourage cross-channel interaction without altering spatial structure. In natural complex signals, real and imaginary components convey complementary information at the same spatial location \citep{imageprocessingbook,complex_survey}, making it important that the generated imaginary features do not introduce substantial new spatial information. Two Phasor Block configurations are used in this work: \phasori performs the initial real-to-complex conversion, while \phasorc further refines complex-valued features (Figure \ref{fig:architecture} (top)). Full architectural diagrams and design details are presented in Appendix \ref{app:phasor_block}.

\paragraph{Psychovisual Framework}\label{sec:psychonet} To experimentally evaluate \ac{DVC} and Phasor Blocks in Section \ref{sec:experiments}, we combine them into a cohesive deep learning framework designed to explicitly mimic psychovisual processing. We emphasize that this framework is not intended as a definitive architecture for achieving state-of-the-art performance, but rather as a proof-of-concept for psychovisually-grounded deep learning. We refer to models utilising this pipeline collectively as PsychoNet.

As illustrated in Figure \ref{fig:architecture}, PsychoNet operates as a coherent dual-domain pipeline comprising the following stages:
\begin{enumerate}
    \item Standard ResNet-style Residual Blocks (ResBlocks) perform initial spatial feature extraction.
    \item Phasor Blocks convert real-valued spatial features into complex-valued representations.
    \item 2D Fast Fourier Transform (FFT) is applied to convert the features from the spatial to the frequency domain. As the magnitude of the DC (zero frequency) and low-frequency components typically dominate visual data, a simple companding operation is also applied to balance the spectrum (see Appendix \ref{app:architecture}).
    \item \ac{DVC} is applied, and its outputs from each frequency band aggregated. For classification, these are further pooled spatially and used for output prediction directly in the frequency domain using a complex-valued linear layer.
\end{enumerate}
Mirroring psychovisual processing, stages 1–2 roughly parallel early cortical feature extraction, while stages 3–4 realise the abstraction step that decouples those features from the high-level representations used for decision-making.


\section{Experiments}\label{sec:experiments}
We evaluate our proposed psychovisual framework using natural image classification as a straightforward, scalable, and well-understood testbed to investigate whether it learns the abstraction-like representations our framing predicts. In the following subsections, we first present visualisations exploring the representations learnt by the core components of PsychoNet. We then use various classification benchmarks to explore its scaling properties and demonstrate it achieves reasonable practical performance, before finally discussing ablation studies.

\paragraph{Implementation details} Our PsychoNet models retain the first two spatial stages of a standard ResNet backbone and replace the remaining deeper stages with our Phasor Blocks and \ac{DVC} module. We evaluate these models across small to large-scale datasets, ranging from the low-resolution CIFAR-10/100 \citep{cifar} (\mytilde50K images) to the moderate-sized ImageNet-100 (\mytilde130K images) and the standard large-scale ImageNet-1K (\mytilde1.2 million images) \citep{imagenet}. We trained both our own and comparison models from scratch using identical recipes; on ImageNet-1K we train for 90 epochs, use an AdamW optimizer with a cosine scheduler and standard augmentations. Full details for all datasets, training, hardware, and model configurations are provided in Appendices \ref{app:classification} and \ref{app:architecture}.


\subsection{Representation analysis}\label{sec:representation_analysis}
In this section, we analyse the internal representations learned by the primary components of our framework. We first investigate what our \ac{DVC} module and Phasor Blocks learn individually, followed by an exploration of their interaction to reveal how \ac{DVC} encodes spatial features extracted by Phasor Blocks. For all visualisations, unless stated otherwise, we utilize a model based on Psycho-B. This model retains the first two ResBlock stages of a ResNet backbone but replaces the remaining stages with Phasor Blocks and an \ac{DVC} module that match the original depth and parameter count.

\begin{figure}[h]
    \centering
    \includegraphics[width=0.9\linewidth]{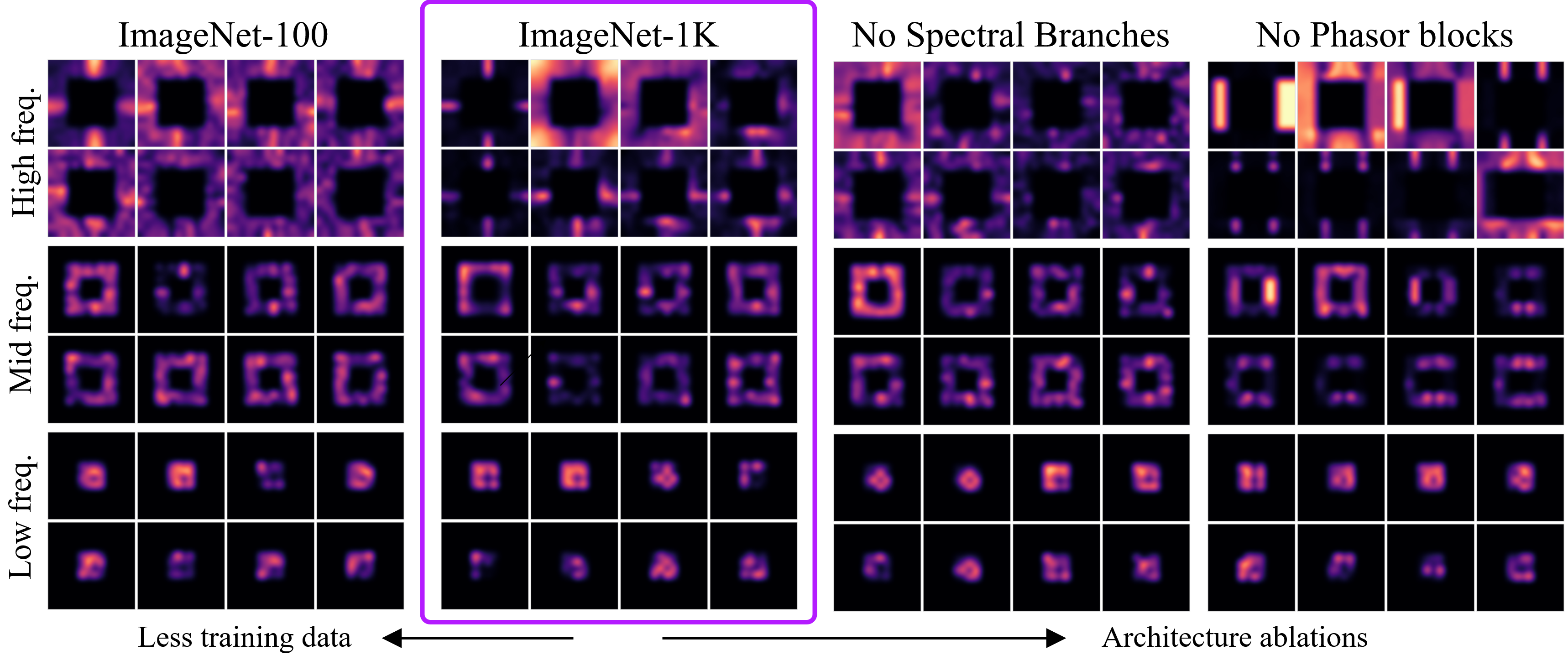}
    \caption{\ac{DVC} filters learnt by Psycho-B trained on ImageNet-100 and ImageNet-1K, as well as for two ablation models on ImageNet-1K. Bilinear smoothing has been applied. `High/mid/low freq.' refer to the [14, 8], [8, 4] and [4, 1] frequency sub-bands created by Spectral Branches. `No Spectral Branches' removes Spectral Branches and uses a single Hadamard Block with global filters - we extract sub-bands only for the visualisation. `No Phasor Blocks' replaces all Phasor Blocks with ResBlocks.}
    \label{fig:pca_filters}
\end{figure}

\paragraph{\ac{DVC} filter learning.} Figure \ref{fig:pca_filters} visualises filters learnt by \ac{DVC}, showing the top spatial principal components as an approximation of the most important frequency features. Remarkably we find that with sufficient training corpora and the full PsychoNet framework, \ac{DVC} learns structured and sparse selections of frequencies that are highly reminiscent of the original hand-crafted psychovisual coding scheme. 
We further investigated how the amount of training data and architectural components affect the quality of these learnt patterns, finding that sparsity requires sufficiently large training corpora: ImageNet-100-trained filters are noticeably noisier than those for ImageNet-1K, and filters trained on CIFAR-10/100 (Figure \ref{fig:filters_datasets}) are noisier still. Additionally, we also find that both Spectral Branches (providing sub-band decomposition) and Phasor Blocks (providing complex-valued features) are important; removing either significantly reduces filter sparsity and expressiveness. Together, these results suggest a naturally emergent psychovisual-style abstraction in the frequency domain, given sufficient data and complex-valued sub-bands. Our further analysis below of their interaction with Phasor Blocks reveals these sparse patterns correspond to structured encodings of meaningful object parts.

\begin{figure}[h]
    \centering
    \includegraphics[width=0.85\linewidth]{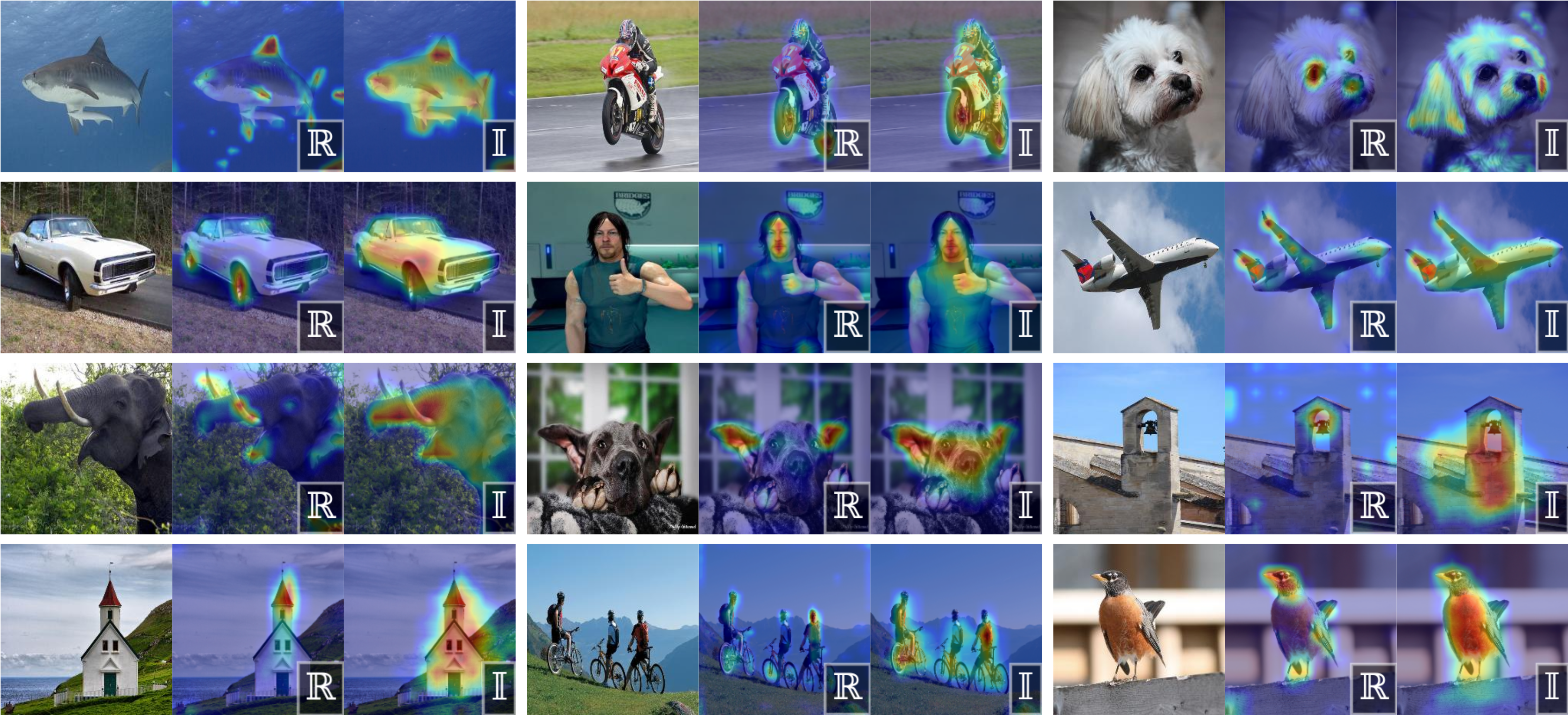}
    \caption{Assorted activation maps (via KPCA-CAM) for mid-level Phasor Blocks of Psycho-B. Real and imaginary components are denoted by $\mathbb{R}$ and $\mathbb{I}$.}
    \label{fig:phasor_kpcacams}
\end{figure}

\paragraph{Phasor Block salience.} We employ KPCA-CAM \cite{kpcacam} for salience mapping to visualise the features extracted by Phasor Blocks. Figure \ref{fig:phasor_kpcacams} shows that Phasor Blocks particularly specialise towards localising characteristic morphological object parts of different classes, such as dog ears, elephant tusks and car wheels.
Since KPCA-CAM only uses activations of the visualised layer and is uninfluenced by model predictions (e.g. via backpropagation in gradient-based CAMs), these results indicate that Phasor Blocks specialise to extract meaningful semantic object parts. This organisation is likely shaped by the presence of \ac{DVC} downstream, which we show below encodes these parts into higher-level abstractions that support interpretable semantic reasoning. Additionally, it appears that the imaginary components of Phasor Block activations capture more global features than the real components (e.g. a dog's face vs. its ears), suggesting a structured and rich utilisation of the complex domain. An initial clustering visualisation of Phasor Block activations is also presented in Figure \ref{fig:clustering}, which finds that observable clustering emerges in both components and becomes increasingly pronounced at deeper layers. These findings further demonstrate that this complex-valued representation is iteratively refined through each block.

\begin{figure}[h]
    \centering
    \includegraphics[width=0.85\linewidth]{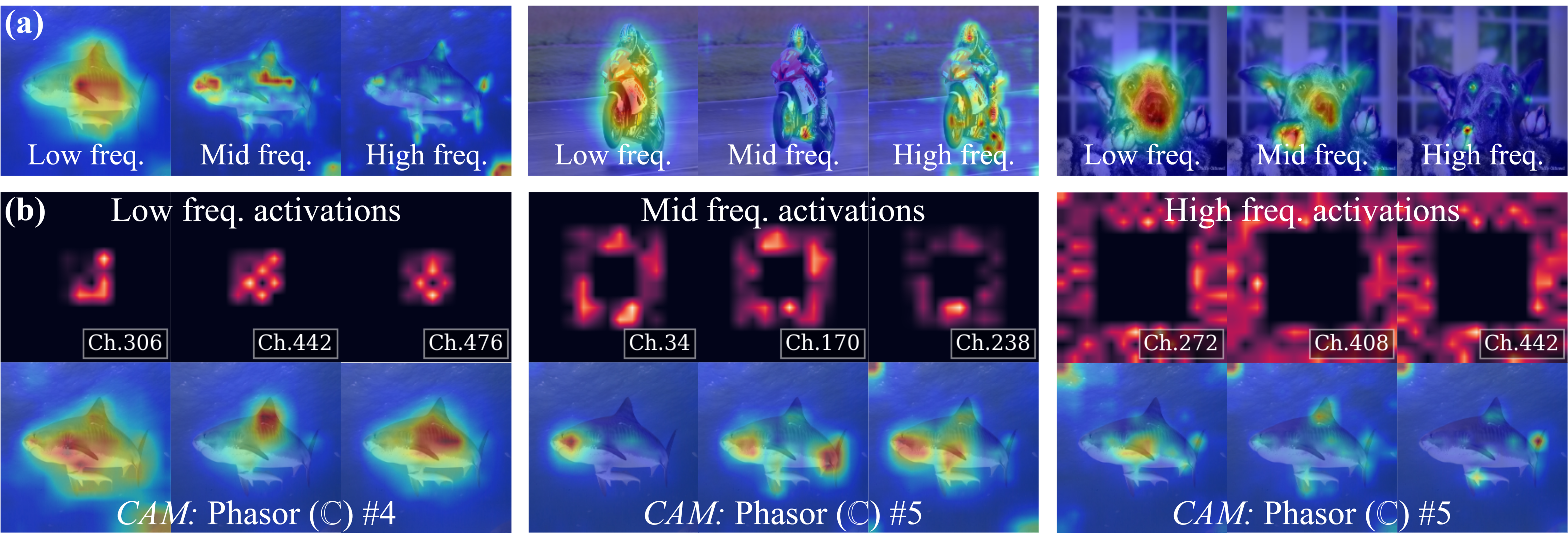}
    \caption{Psycho-B Phasor Block salience maps (via HiResCAM) conditioned on gradients \textbf{(a)} from individual Spectral Branch sub-bands and \textbf{(b)} from individual frequency domain feature channels.}
    \label{fig:grad_cams}
\end{figure}

\paragraph{\ac{DVC} and Phasor Block interaction.} We use a modified version of HiResCAM \citep{hirescam} to explore how \ac{DVC} uses the semantic features produced by Phasor Blocks. Normal HiResCAM produces salience maps by element-wise multiplying layer activations with gradients backpropogated from model predictions, so the salience regions have a high contribution to the final classification prediction. We extend this approach to isolate regions used by specific branches and filters of \ac{DVC} by selectively masking (setting to zero) gradient contributions from the other components. First, we examine each of the three sub-bands created by PsychoNet's Spectral Branches. After masking gradients of Hadamard Blocks for all but one sub-band, Phasor Blocks' salience regions reveal that \ac{DVC} distributes object parts by scale. Figure \ref{fig:grad_cams} (a) shows that the low-frequency sub-band focuses on subjects broadly, while mid-high frequencies isolate more specific parts of different sizes. This aligns with frequency domain theory, in which low frequencies capture coarse spatial structure and higher frequencies finer detail and edges, supporting the view that \ac{DVC} performs structured filtering in the frequency domain. We also isolate activations from individual Hadamard Block channels, showing that within each band, channels specialise to distinct object parts and correspond to distinct sparse frequency selections (Figure \ref{fig:grad_cams} (b)).

Overall, these result suggest that \ac{DVC} learns a semantic intermediate representation that encodes selections of object parts. Given that \ac{DVC} is placed immediately before the decision making (classification) layers of PsychoNet, it is likely selecting those most relevant to the task. In doing so, \ac{DVC} functions as an abstraction bridging part extraction in Phasor Blocks and higher-level semantic reasoning, mirroring the role of abstractions used in psychovisual processing.

\subsection{Quantitative evaluation}
Unlike ResNet architectures which rely on increasing depth to scale representational capacity \citep{resnet}, we hypothesize that because PsychoNet’s high-level processing is handled by frequency domain modules, it should remain relatively depth-independent. To test this, we designed Psycho-S/B/L/H to match four ResNet parameter sizes while sharing the same early spatial layers; however, rather than adding depth, we scale representational capacity by increasing the channel width of existing Phasor Blocks and \ac{DVC} filters. As the base PsychoNet models only downsample features to $14\times14$ resolution at the smallest in order to maintain spectral fidelity for \ac{DVC}, they use higher FLOPs than ResNet which downsamples further to $7\times7$. As such, we also created three efficient PsychoNet variants, Psycho-Eff-S/B/L, which do downsample to $7\times7$ and extract frequency features in the 7 to 14 sub-band from an earlier layer. These have similar FLOPs with ResNet at the cost of a small performance compromise compared to the normal PsychoNet models. Table \ref{tab:results_main_body} summarises the results from all classification experiments.
\begin{figure}[h]
    \centering
    \includegraphics[width=0.98\linewidth]{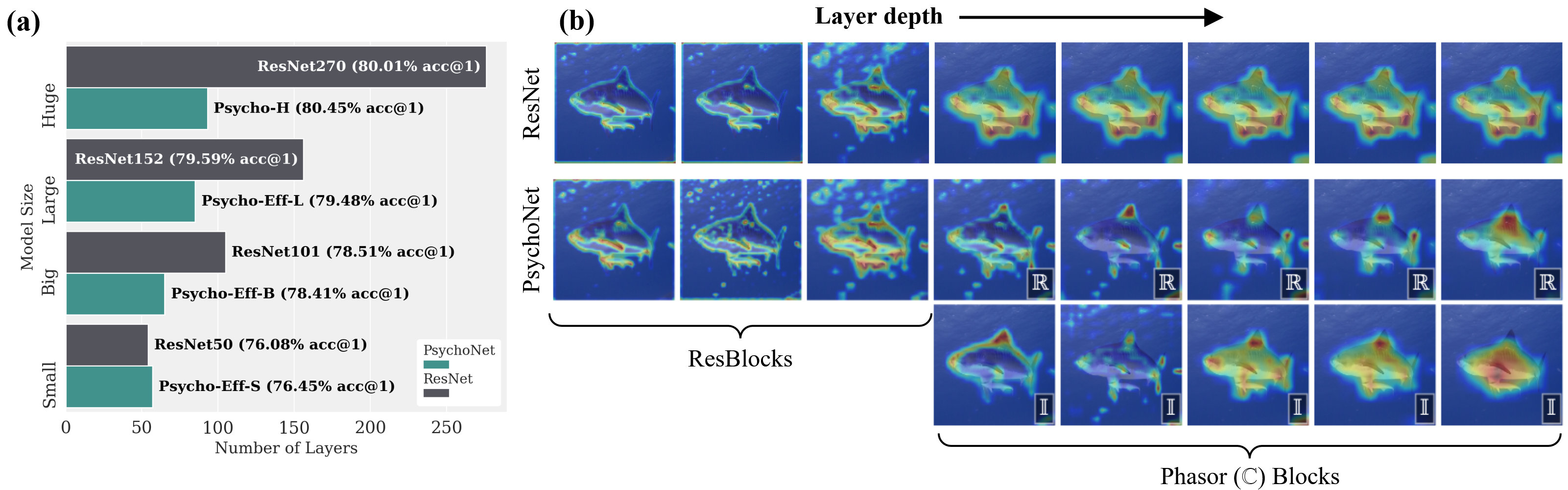}
    \caption{\textbf{(a)} Comparison of layer depth between different ResNet models and our shallowest PsychoNet model of comparable size. \textbf{(b)} Comparison between activation maps (via KPCA-CAM) of Psycho-B and ResNet101 for a range of layer depths. Real and imaginary components are denoted by $\mathbb{R}$ and $\mathbb{I}$.}
    \label{fig:psycho_vs_resnet_results}
\end{figure}

\begin{table}[ht]
    \centering
    \caption{Summary of classification results (\% top-1 accuracies) on CIFAR-10, CIFAR-100, ImageNet-100 (IN100) and ImageNet-1K (IN1K). Each pair of rows (separated by horizontal lines) compares a baseline \ac{CNN} to PsychoNet model(s) of comparable size.}
    \begin{tabularx}{1\textwidth}{lllrrrrr}
        \toprule
        Model & Param. (M) & Layers & GFLOPs & CIFAR-10 & CIFAR-100 & IN100 & IN1K\\
        \midrule
    
        ResNet50 & 25.56 & 54 & 8.18 & 94.14 & 78.10 & 80.90 & 76.04\\
        \rowcolor{gray!20} Psycho-S & 25.35 & 65 & 12.31 & \textbf{95.08} & \textbf{78.97} & 82.50 & \textbf{76.86}\\
        \rowcolor{gray!20} Psycho-Eff-S & 26.27 & 57 & 8.01 & 94.86 & 78.15 & \textbf{83.20} & 76.45 \\
        \midrule
    
        ResNet101 & 44.55 & 105 & 15.60 & 93.64 & 79.13 & 81.90 & 78.43\\
        \rowcolor{gray!20} Psycho-B & 42.01 & 93 & 30.13 & \textbf{94.99} & \textbf{79.49} & 83.60 & \textbf{78.85}\\
        \rowcolor{gray!20} Psycho-Eff-B & 45.82 & 65 & 15.80 & 94.98 & 78.22 & \textbf{83.92} & 78.41 \\
        \midrule
    
        ResNet152 & 60.10 & 156 & 23.03 & 93.17 & 77.51 & 83.60 & 79.59\\
        \rowcolor{gray!20} Psycho-L & 61.28 & 93 & 54.47 & \textbf{94.95} & \textbf{79.64} & 84.82 & \textbf{79.85}\\
        \rowcolor{gray!20} Psycho-Eff-L & 62.03 & 85 & 23.06 & \textbf{94.95} & 78.18 & \textbf{84.98} & 79.48 \\
        \midrule
    
        ResNet270 & 89.60 & 276 & 40.50 & 76.51 & 50.87 & 83.80 & 80.01\\
        \rowcolor{gray!20} Psycho-H & 88.61 & 93 & 64.12 & \textbf{94.68} & \textbf{79.89} & \textbf{85.00} & \textbf{80.45} \\
        \midrule
    
        ConvNeXt-S & 50.22 & 113 & 17.36 & 94.09 & 76.96 & \textbf{86.98} & \textbf{80.78}\\
        \rowcolor{gray!20} PsychoDW & 49.51 & 106 & 27.42 & \textbf{95.46} & \textbf{79.67} & 86.76 & 80.59\\
        \bottomrule
    \end{tabularx}
    \label{tab:results_main_body}
\end{table}

Overall, PsychoNet demonstrates significantly less dependence on depth for scaling than traditional ResNet architectures, as illustrated in Figure~\ref{fig:psycho_vs_resnet_results}(a). This is evidenced by Psycho-Eff-B and Psycho-Eff-L, which use 38\% and 45.5\% fewer layers than ResNet-101 and ResNet-152 while closely matching their parameters, FLOP counts, and performance. Similarly, Psycho-L and Psycho-H utilize \mytilde1.7$\times$ and \mytilde3$\times$ fewer layers than ResNet-152 and ResNet-270 respectively, while achieving improved accuracy across every dataset. This demonstrates that \ac{DVC} is able to subsume the role of a significant portion of deep spatial layers, showing it is an effective high-level processing module. This aligns with the psychovisual framing that motivated our design: \ac{DVC} operates on globally-focused frequency-domain abstractions, whereas spatial features are sparse and locally structured, requiring significant depth to progressively aggregate them. Furthermore, PsychoDW achieves performance comparable to ConvNeXt-S, demonstrating that PsychoNet can also extend to match more modern architectures. Figure~\ref{fig:psycho_vs_resnet_results}(b) provides complementary qualitative evidence for this: unlike ResNet, which produces diffuse, unstructured activations throughout its depth, PsychoNet's spatial layers consistently localise distinct morphological object parts, suggesting that Phasor Blocks and \ac{DVC} together successfully take over the high-level processing otherwise requiring many additional spatial layers.

\subsection{Ablation studies}
We conduct two ablation studies to quantify the contribution of PsychoNet's key components: (1) minimal CIFAR-10 models ($\sim$3M parameters) examining the isolated effects of \ac{DVC} and Phasor Blocks (Table~\ref{tab:barebones_ablation}), and (2) Psycho-L on ImageNet-1K, targeting the components that our qualitative results suggested are particularly important at scale — Spectral Branching, which visibly improved \ac{DVC} filter sparsity and expressivity, and full Phasor Blocks, which produced compelling interpretable salience maps on large-scale data (Table~\ref{tab:imagenet_ablation}).

\begin{table}[ht]
    \centering
    \begin{minipage}[t]{0.48\textwidth}
        \centering
        \captionof{table}{\textbf{Barebones architecture ablation} on CIFAR-10.}
        \label{tab:barebones_ablation}
        \begin{tabularx}{\textwidth}{lXXX}
            \toprule
            Model & \ac{DVC} & Phasor Blocks & Top-1 Acc. (\%) \\
            \midrule
            I   & \ding{51} & \ding{51} & 94.03 \\
            II  & \ding{55} & \ding{51} & 92.86 \\
            III & \ding{51} & \ding{55} & 94.20 \\
            IV  & \ding{55} & \ding{55} & 93.33 \\
            \bottomrule
        \end{tabularx}
    \end{minipage}
    \hfill
    \begin{minipage}[t]{0.48\textwidth}
        \centering
        \captionof{table}{\textbf{ImageNet-1K ablation} on Psycho-L (Model A) and variants.}
        \label{tab:imagenet_ablation}
        \begin{tabularx}{\textwidth}{lXXX}
            \toprule
            Model & Spectral Branch & Phasor Blocks & Top-1 Acc. (\%) \\
            \midrule
            A & \ding{51} & \ding{51} & 79.85 \\
            B            & \ding{55} & \ding{51} & 79.27 \\
            C            & \ding{51} & \ding{55} & 79.60 \\
            D            & \ding{55} & \ding{55} & 79.12 \\
            \bottomrule
        \end{tabularx}
    \end{minipage}
\end{table}

For the minimal CIFAR-10 ablation, models removing \ac{DVC} pass the frequency domain features after \ac{FFT} directly into the classification layers, while models removing Phasor Blocks replace them with basic Conv. Blocks (see Appendix \ref{app:ablation}), maintaining approximate parameter count and model depth. Table~\ref{tab:barebones_ablation} shows that removing \ac{DVC} drops performance by 1.17\% (Model I vs.\ II) and 0.87\% (Model III vs.\ IV), highlighting its contribution to the framework. However, at this small scale, substituting Phasor Blocks for Conv. Blocks actually slightly improves classification accuracy, perhaps as Phasor Blocks are overly complex relative to the task.

\begin{figure}[ht]
    \centering
    \includegraphics[width=0.7\linewidth]{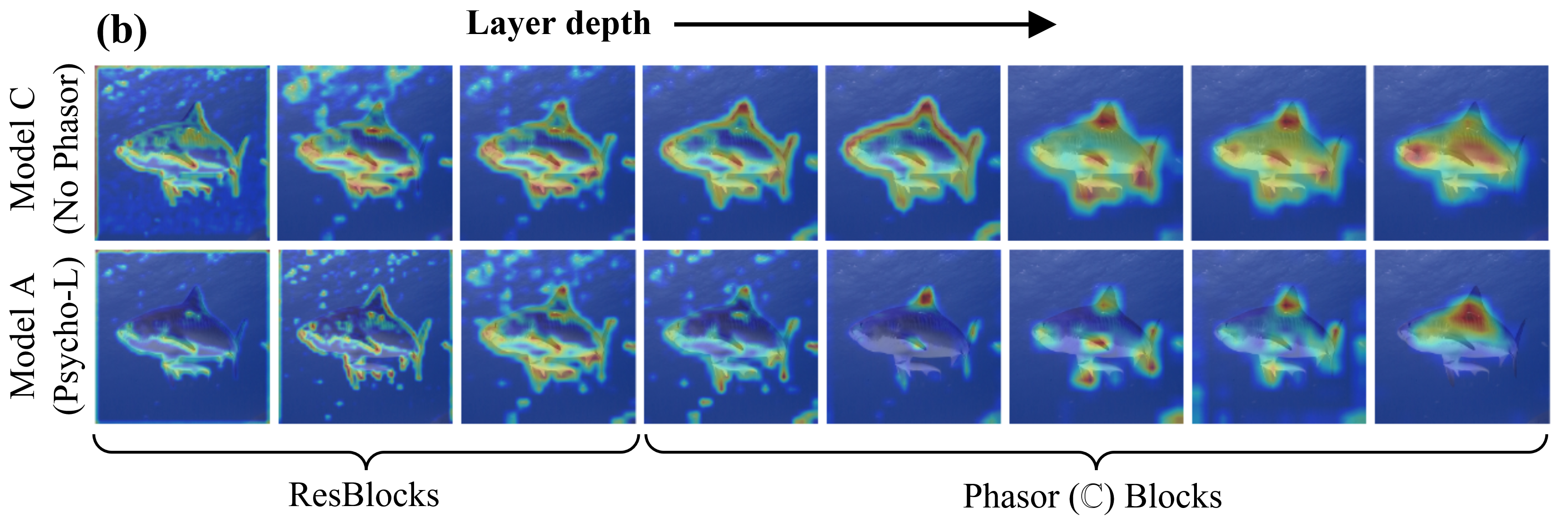}
    \caption{Comparison between activation maps of Ablation Model A and C.}
    \label{fig:cams_ablation}
\end{figure}

For the ImageNet-1K ablation, we evaluate variations of the Psycho-L architecture (Table~\ref{tab:imagenet_ablation}). Models without Spectral Branches replace the multiple branches in \ac{DVC} with a single global filter lacking DropCrop operations. Models without Phasor Blocks replace all but one final Phasor ($\mathbb{I}$) block with ResBlocks while maintaining parameter size. We observe that removing Spectral Branches leads to notable accuracy drops of 0.58\% (A vs.\ B) and 0.47\% (C vs.\ D), showing that this spectral decomposition not only improves filter sparsity but also makes a meaningful impact on performance. The removal of Phasor Blocks results in smaller drops of 0.25\% (A vs.\ C) and 0.14\% (B vs.\ D). However, consistent with our earlier qualitative findings, the value of Phasor Blocks lies not in classification accuracy but in representational quality: as shown in Figure~\ref{fig:cams_ablation}, Model~A produces activation maps capturing meaningful object parts, whereas Model~C yields comparatively diffuse activations lacking clear structure. Together, these ablations confirm distinct roles within the psychovisual pipeline: Spectral Branches structure \ac{DVC}'s abstractions, while Phasor Blocks preserve the part-level features feeding them.

\section{Conclusion}
In this work, we proposed \acl{DVC}, a data-driven adaptation of psychovisual coding. Integrated into our PsychoNet framework, \ac{DVC} naturally learns sparse coronal frequency patterns that closely resemble classical hand-crafted codes, suggesting these are a natural representation scheme for psychovisual abstraction. Unlike conventional \acp{CNN}, PsychoNet enforces a clear structural separation of stages: Phasor Blocks isolate semantically meaningful object parts, which \ac{DVC} then organizes into interpretable, frequency domain representations for decision-making. Our results further show that \ac{DVC} subsumes the high-level processing role of deep spatial layers, matching or exceeding ResNet baselines with significantly fewer layers. Together, these findings provide strong evidence that psychovisual coding is a principled and viable basis for interpretable representation learning, with promising implications for building models whose internal reasoning is transparent by design.

\paragraph{Limitations and Future Work.}
While we were able to show \ac{DVC} organises and encodes selections of meaningful object parts, further work is still required to determine how the deeper semantic meaning of these abstractions should be interpreted in relation to broader notions of reasoning, such as those studied in neuroscience \citep{brain_representations_1, brain_representations_2}. We also aim to apply PsychoNet beyond classification, particularly to dense prediction tasks which may allow \ac{DVC} to utilise wider frequency ranges. Finally, it is also known that aliasing can afflict standard \ac{CNN} architectures \citep{cnn_aliasing}; future work should assess its impact on our frequency domain representations and whether mitigation can improve results.

\newpage
\small
\bibliography{references}
\bibliographystyle{unsrtnat}

\newpage
\appendix

\newpage
\appendix
\setcounter{figure}{0}
\setcounter{table}{0}
\renewcommand{\thefigure}{A.\arabic{figure}}
\renewcommand{\thetable}{A.\arabic{table}}
\begin{center}
    \huge\textbf{Appendices}
\end{center}

In the following we present appendices to our work, structured as follows: Appendix \textcolor{orange}{\ref{app:abbrev}} lists the abbreviations used throughout our work. Appendix \textcolor{orange}{\ref{app:background}} presents additional background material. Appendix \textcolor{orange}{\ref{app:classification}} presents dataset details and training recipes for our classification experiments. Appendix \textcolor{orange}{\ref{app:architecture}} provides detailed information about the architectural configurations of all models used in our main classification experiments. Appendix \textcolor{orange}{\ref{app:ablation}} provides architecture and training details for our ablation studies.

\section{Abbreviations}\label{app:abbrev}
\begin{acronym}[1D]
\acro{1D}{one dimensional}
\acro{2D}{two dimensional}
\acro{3D}{three dimensional}
\acro{DVC}{Deep Visual Coding}
\acro{CNN}{Convolutional Neural Network}
\acro{FT}{Fourier Transform}
\acro{DFT}{Discrete Fourier Transform}
\acro{IDFT}{Inverse Discrete Fourier Transform}
\acro{FFT}{Fast Fourier Transform}
\acro{IFFT}{Inverse Fast Fourier Transform}
\acro{MRI}{Magnetic Resonance Imaging}
\acro{RF}{Receptive Field}
\acro{PC}{Principle Component}
\end{acronym}

\section{Background}\label{app:background}
In this section we present additional background and details about psychovisual coding, the Fourier Transform and complex-valued networks.

\subsection{Psychovisual Coding}\label{app:psychovisual}
\begin{figure}[h]
    \centering
    \includegraphics[width=0.4\linewidth]{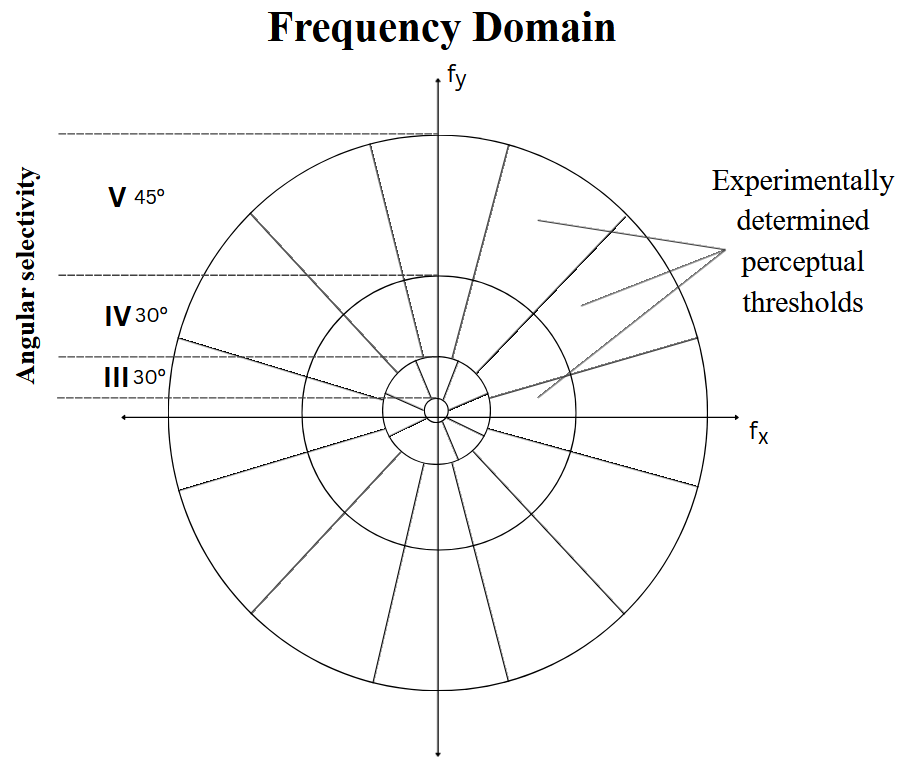}
    \caption{Coronal frequency sub-bands used in psychovisual coding from \citet{saadane_visual_1998}.}
    \label{fig:psychovisual_coding}
\end{figure}

Our work is inspired by groundbreaking research conducted in the 1990s by French researchers led by Dominique Barba in understanding the human aspect of mammalian vision, i.e. the psychovisual capability of the human brain for visual perception arising from the need for early television signal compression~\citep{hanen_high-quality_1993}. At the time, statistical approaches based on Shannon's information theory and rectilinear methods such as discrete (Haar) wavelets were popular, and they argued that these approaches were sub-optimal because they treated all errors equally. They would propose psychovisual quantizers as an efficient form of image coding that would retain the important image information pertaining to its interpretation by the human vision system and quantization matched the detection thresholds of the visual cortex~\citep{senane_image_1995}. These quantizers were proposed to be the coronas of the 2D Fourier space, where the model of the vision system assumes Fourier space is analyzed using radial symmetric functions~\citep{saadane_design_1994, saadane_visual_1998} (see Figure~\ref{fig:psychovisual_coding}), which they showed can also be mapped to colors in human vision~\citep{callet_interactions_1999}. The premise is that visual recognition and feature extraction could be performed by selecting coronal sectors of Fourier space directly through the quantisation of adjacent frequencies, thereby providing directional band limited filtering within the scene. Their psychophysical experiments were also used to select the optimal sub-bands that allowed image compression that was difficult for humans to distinguish~\citep{saadane_masking_2001}. Our \ac{DVC} is a data-driven adaptation of this approach, using band-limited frequency filters to learn sparse frequency selections using supervisory signals from classification and segmentation tasks.

This work on visual codes over the course of a decade would result in among the first uses of vector quantization for image coding~\citep{senane_image_1995}, a perceptually based image quality metric~\citep{saadane_masking_2001} and one of the foundations of discrete projection theory, where a central slice theorem is established for discrete Fourier space based as exact 1D forms of these psychovisual radial functions as slices and therefore projections in image space~\citep{guedon_psychovisual_1995}. This work would even pioneer the use of the wavelet transform to projection data before it would be formalized as ridgelets by Candes and Donoho~\citep{candes_ridgelets:_1999}. The Mojette transform would itself form the basis of an entire area of discrete tomography that creates discrete projections of images~\citep{normand_controlled_1996} in diverse areas such as image reconstruction~\citep{kingston_generalised_2007, chandra_robust_2014} and compression~\citep{guedon_load-balancing_2001}, computed tomography~\citep{hou_parallel-beam_2013} and network transmission~\citep{parrein_scalable_2001, verbert_distributed_2002}. Although the number of publications is too numerous to list here, a summary of these works and areas can be found in the Mojette transform book~\citep{guedon_mojette_2013}.


\subsection{The Frequency Domain}\label{app:background_freq}
Intuitively, \acf{DFT} is the projection of any (finite) signal on the set of (harmonic) orthonormal basis functions created via the unit circle as the $n$th root of unity $e^{2\pi i/N} = 1$ with $N\in\mathbb{Z}$ and $i^2=-1$~\citep{cooley_finite_1969}, i.e., the cyclotomy or division of a unit circle into equal parts. This projection results in weights for each of these harmonics known as the Fourier coefficients required to represent the signal and comprise in its frequency domain representation. Formally, the \ac{DFT} $\ \hat{I}(\FFTfirstIndex,\FFTsecondIndex)$ of an $N\times N$ image $I(\firstIndex,\secondIndex)$ is defined as 
\begin{equation}
 \hat{I}(\FFTfirstIndex,\FFTsecondIndex) = \sum\limits^{N-1}_{\firstIndex = 0} \sum\limits^{N-1}_{\secondIndex = 0} I(\firstIndex,\secondIndex) e^{-2\pi i\FFTfirstIndex\firstIndex/N} e^{-2\pi i\FFTsecondIndex\secondIndex/N}.
\end{equation}
The \acf{IDFT} is defined as
\begin{equation}
  I(\firstIndex,\secondIndex) = \frac{1}{N^2} \sum\limits^{N-1}_{\FFTfirstIndex = 0} \sum\limits^{N-1}_{\FFTsecondIndex = 0} \hat{I}(\FFTfirstIndex,\FFTsecondIndex) e^{2\pi i\FFTfirstIndex\firstIndex/N} e^{2\pi i\FFTsecondIndex\secondIndex/N}.
\end{equation}
Both the \ac{DFT} and \ac{IDFT} are row/column separable. 

This \ac{DFT} space has many useful properties, the main being the transformation of convolution operations into element-wise multiplication (i.e. the Hadamard product) known as the circular convolution property. Let $x[j, k], y[j, k], j, k \in 0, ..., N-1$ be two discrete $N \times N$ spatial signals. The circular convolution of these two signals is defined as  
\begin{equation}
    x[j, k] * y[j, k] = \frac{1}{N^2} \sum_{m=0}^{N-1} \sum_{n=0}^{N-1} x[m, n] y[\langle(j - m)\rangle_N, \langle(k - n)\rangle_N],
\end{equation}
where $\langle\cdot\rangle_N$ denotes modulo $N$. The Convolution Theorem \citep{imageprocessingbook} then states that:
\begin{equation}
    \mathcal{F}\sqr{x * y} = \mathcal{F}\sqr{x} \odot \mathcal{F}\sqr{y} \text{    or equivalently    } x * y = \mathcal{F}^{-1} \sqr{\mathcal{F}\sqr{x} \odot \mathcal{F}\sqr{y}}
    \label{eq:convolution_theorem}
\end{equation}
where $\mathcal{F}[.]$ and $\mathcal{F}^{-1}[.]$ denote the \ac{DFT} and \ac{IDFT}, and $\odot$ the Hadamard product. Hence, circular convolution in the spatial domain is equivalent to applying the Hadamard product in the frequency domain. As such, the frequency domain is highly conducive to global representations, since each element of an image's frequency spectra presents a unique global view of the image, analogous to convolving it with a directional striped kernel. In practice, the \ac{DFT} and \ac{IDFT} are computed using the Fast Fourier Transform and Inverse Fast Fourier Transform respectively \citep{cooley_finite_1969}. 

\subsection{Complex-valued Neural Networks}\label{app:background_complex}
Most work for complex-valued neural networks involve developing components of these networks to work in the complex domain, such as activation functions \citep{scardapane_complex-valued_2018}. Most complex-valued \ac{CNN}s use the network blocks introduced by \citet{deep_complex}. The distributive property of convolution allows convolution between a complex input $\bm{h} = \bm{a} + i\bm{b}$ and a complex kernel $\bm{W} = \bm{W_R} + i\bm{W_I}$ to be decomposed into four real-valued component wise convolutions:

\begin{equation}
        \bm{W} * \bm{h} = \left(\bm{W_R} *\bm{a} - \bm{W_I}*\bm{b}\right) + i\left(\bm{W_I}*\bm{a} + \bm{W_R} *\bm{b}\right)
\end{equation}

Consequently, complex-valued convolution layers are usually more computationally and memory intensive (additionally stores imaginary features) than real-valued ones. \citet{deep_complex} also developed complex normalization methods and activation functions. Complex-valued modules in PsychoNet use the complex-valued convolution ($\mathbb{C}$Conv) and batch-normalization ($\mathbb{C}$BN) layers from \citet{deep_complex}, and a naïve adaptation of the GELU activation function ($\mathbb{C}$GELU) which just applies the original function to real and imaginary channels separately.

When applying complex-valued networks to real-valued images, most works use a small initial module to convert the input into complex-valued features. However, such approaches have yielded only minor improvements in the past over directly using real-valued networks \citep{deep_complex,complex_densenet}. Accordingly, recent complex-valued networks predominantly focus on domains with naturally complex data, such as \ac{MRI}, radar and audio signal processing \citep{complex_unet,covegan,analysisofcomplexnetworks,complex_survey,deep_complex}. To try bridge this gap, a complex-valued colour space by reinterpreting the cylindrical coordinates of the HSV colour model as 2D magnitude and phase was developed \citep{fccn}. They applied this to standard complex-valued CNNs, improving results on common image classification tasks, but retained the high complexity of complex-valued networks. On the other hand, PsychoNet primarily uses real-valued modules (Phasor Blocks) that learn to generate \textit{complementary} complex-valued features to given real features, as described in Section \ref{sec:method}.

\section{Experiment setup}\label{app:classification}
In this section we present complete dataset details, training recipes and some additional results for the classification experiments conducted. 

\subsection{ImageNet-1K}
We use the standard large ImageNet-1K subset from \citep{imagenet} containing \mytilde1.2 million training and \mytilde50000 images for validation/testing. Table \ref{tab:in1k_recipe} presents the training recipe used for ImageNet experiments.  

\begin{table}[ht]
    \caption{ImageNet training recipe}
    \label{tab:in1k_recipe}\textbf{}
    \centering
    \footnotesize
    \begin{tabularx}{0.85\textwidth}{l@{\hspace{3em}}X}
        \toprule
        Setting & Value \\
        \midrule
        Image size & $224\times 224$\\
        Epochs & 90 \\
        \makecell[l]{Batch size\\ \quad (overall, not per GPU)} & 1024 \\
        \midrule

        Loss & Cross entropy \\
        Optimizer & AdamW ($\beta_1=0.9, \beta_2=0.999$) \\
        Scheduler & cosine \\
        Initial learning rate (LR) & $5 \cdot 10^{-4}$\\
        Warmup & warmup LR $=10^{-6}$ , 5 epochs\\
        Learning rate decay & min. LR $=10^{-5}$, 12 epochs\\
        \midrule

        Augmentation &
        \makecell[l]{
            resize, crop, interpolate, horizontal flip, RandAugment,\\
            \quad \quad MixUp, CutMix, label smoothing
        }\\
        \midrule
        
        GPU &
         \makecell[l]{
            $2 \times$ NVIDIA H100: Psycho-B, ResNet101,\\
            \quad \quad all `Big' sized ablation models \\
            $2 \times$ AMD MI300X: Psycho-S, ResNet50 \\
            $4 \times$ AMD MI300X: All other models
         }\\
         \bottomrule
    \end{tabularx}
\end{table}

\begin{figure}[ht]
    \centering
    \includegraphics[width=0.85\linewidth]{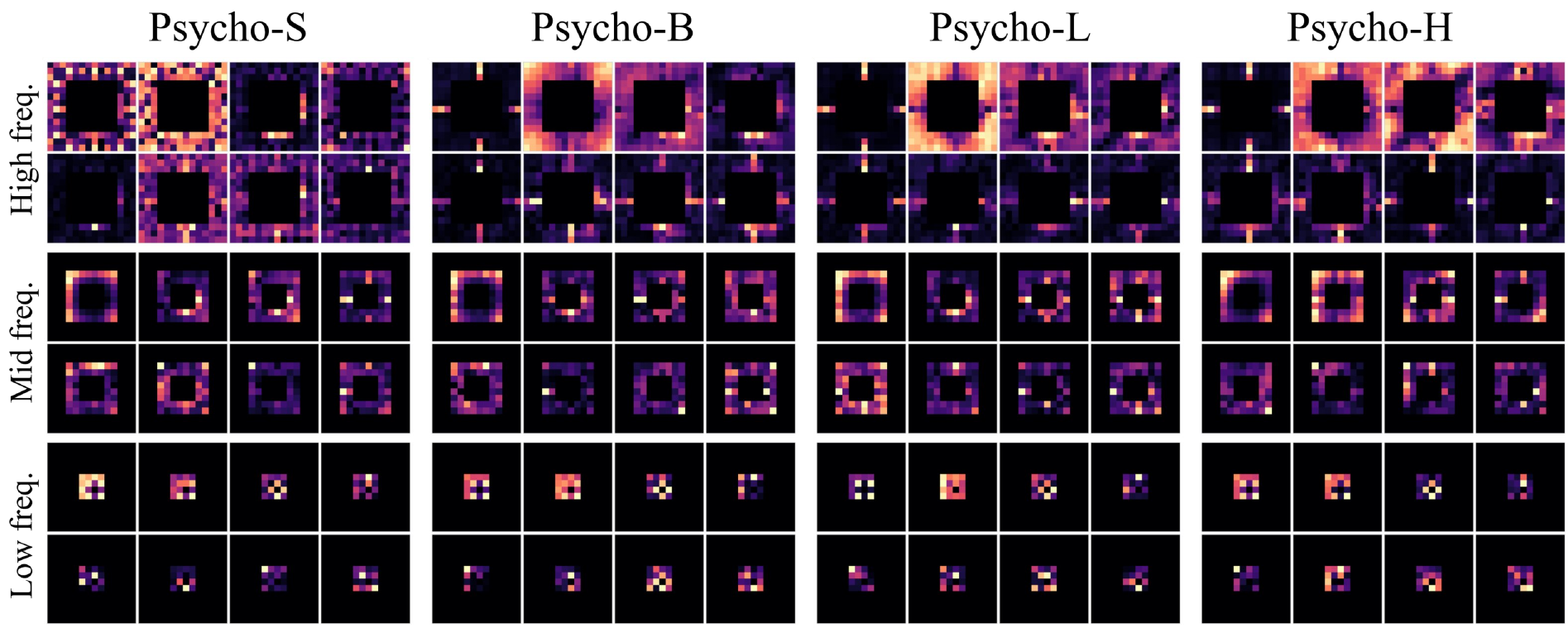}
    \caption{Top \acp{PC} of \ac{DVC} filters learnt by different sized ResNet-based PsychoNet models on ImageNet-1K. `High/mid/low freq.' refer to the [14, 8], [8, 4] and [4, 1] frequency sub-bands created by Spectral Branches.}
    \label{fig:filters_model_size}
\end{figure}

\begin{figure}[ht]
    \centering
    \includegraphics[width=0.85\linewidth]{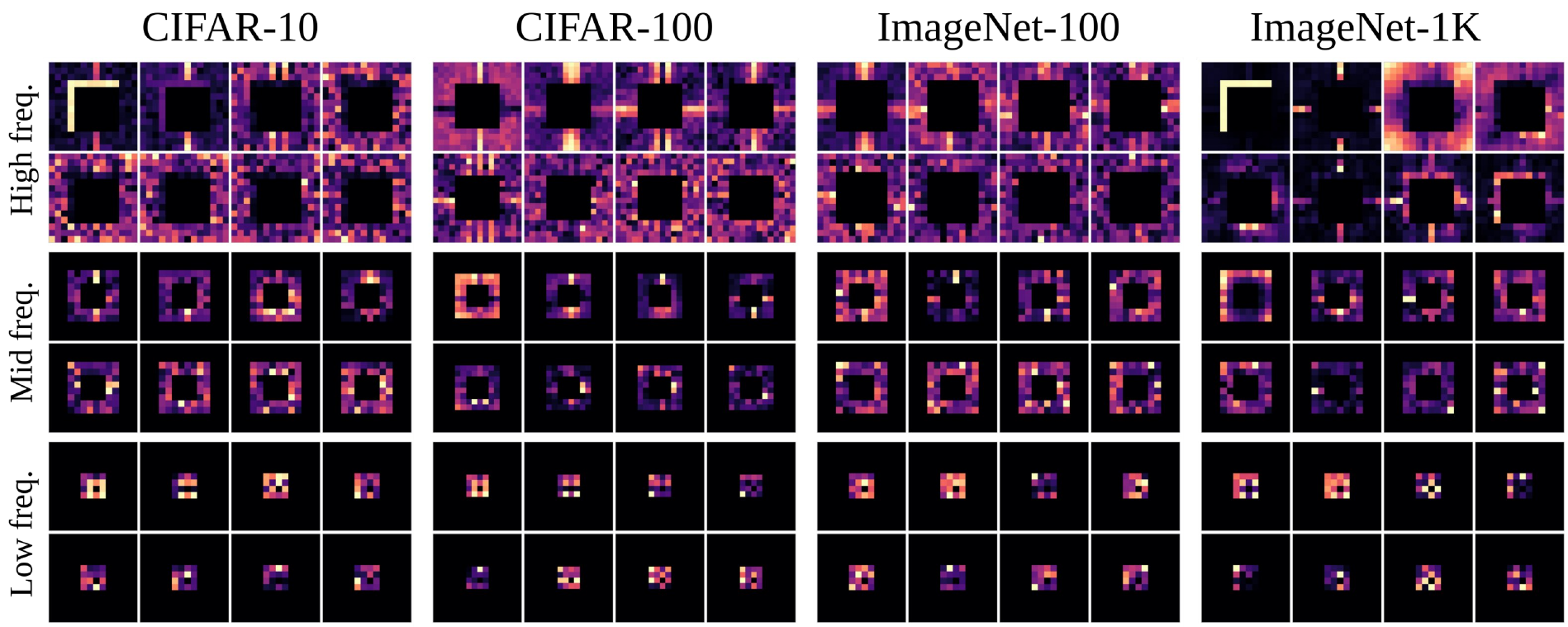}
    \caption{Top \acp{PC} of \ac{DVC} filters learnt by Psycho-B on different resolution and size datasets. `High/mid/low freq.' refer to the [14, 8], [8, 4] and [4, 1] frequency sub-bands created by Spectral Branches.}
    \label{fig:filters_datasets}
\end{figure}

Table \ref{tab:in1k_recipe} presents all ImageNet-1K experiment results. PsychoNet moderately improves top-1 accuracy for all ResNet baselines ($\uparrow$ 0.82\%, 0.41\%, 0.26\% and 0.44\% vs. ResNet50 to 270), and incurs a small decrease for ConvNeXt-S ($\downarrow$ 0.19). Figure \ref{fig:filters_model_size} compares \ac{DVC} filters learnt by different ResNet-based PsychoNet sizes, showing that with larger model size, the filters become increasingly structured and sparser, with clearer frequency selectivity and reduced noise. Figure \ref{fig:filters_datasets} compares \ac{DVC} filters learnt by Psycho-B on ImageNet-1K to the smaller resolution/size datasets in Appendix \ref{sec:app_small_classification}. It is evident that increasing image resolution and dataset size both yield much sparser filters. These results suggest that the sparse patterns correspond to a data-driven representation naturally emergent from visual information.


\subsubsection{PsychoDW representation analysis}
\begin{figure}[ht]
    \centering
    \includegraphics[width=0.98\linewidth]{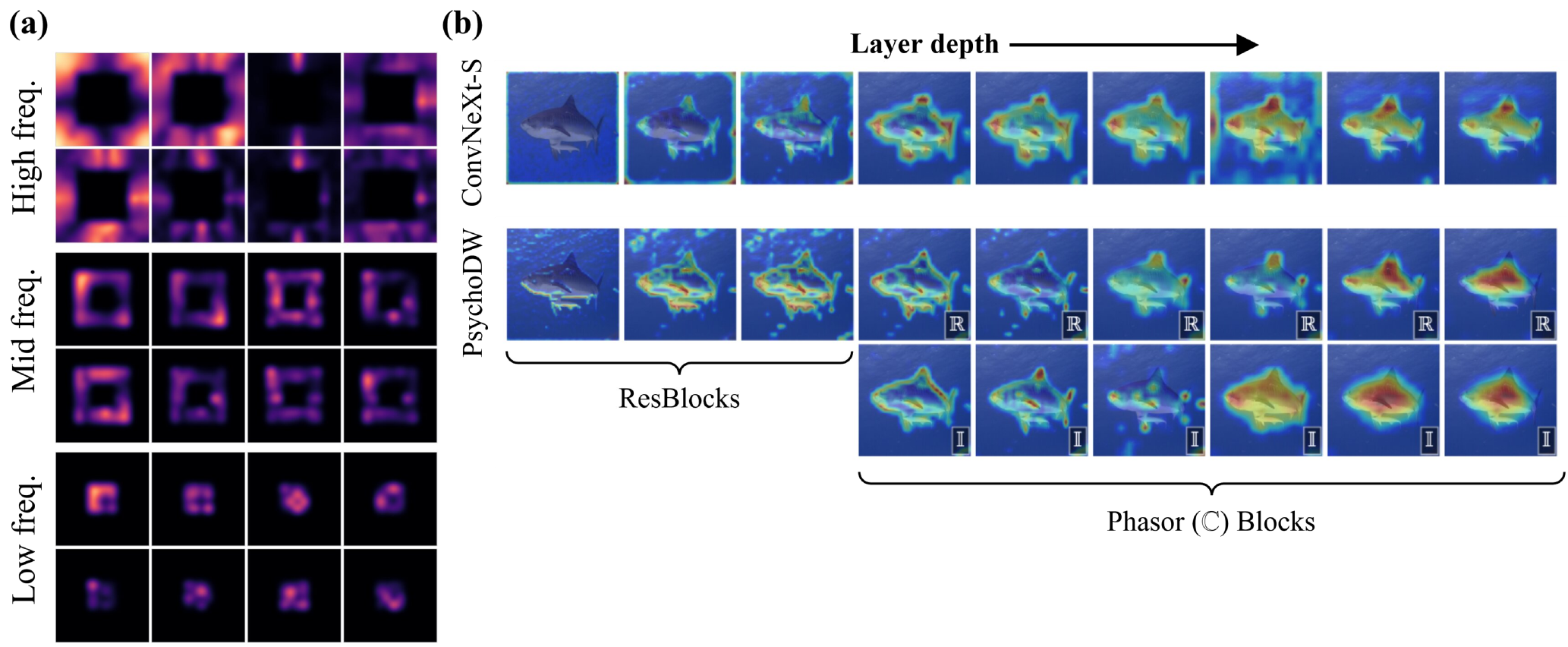}
    \caption{\textbf{(a)} Top spatial principal components of \ac{DVC} filters learnt by PsychoDW trained on ImageNet-1K. Bilinear smoothing has been applied. \textbf{(b)} Comparison between activation maps (via KPCA-CAM) of PsychoDW and ConvNeXt-S for a range of layer depths. Real and imaginary components are denoted by $\mathbb{R}$ and $\mathbb{I}$.}
    \label{fig:psychodw_vs_convnext}
\end{figure}

\begin{figure}[ht]
    \centering
    \includegraphics[width=0.98\linewidth]{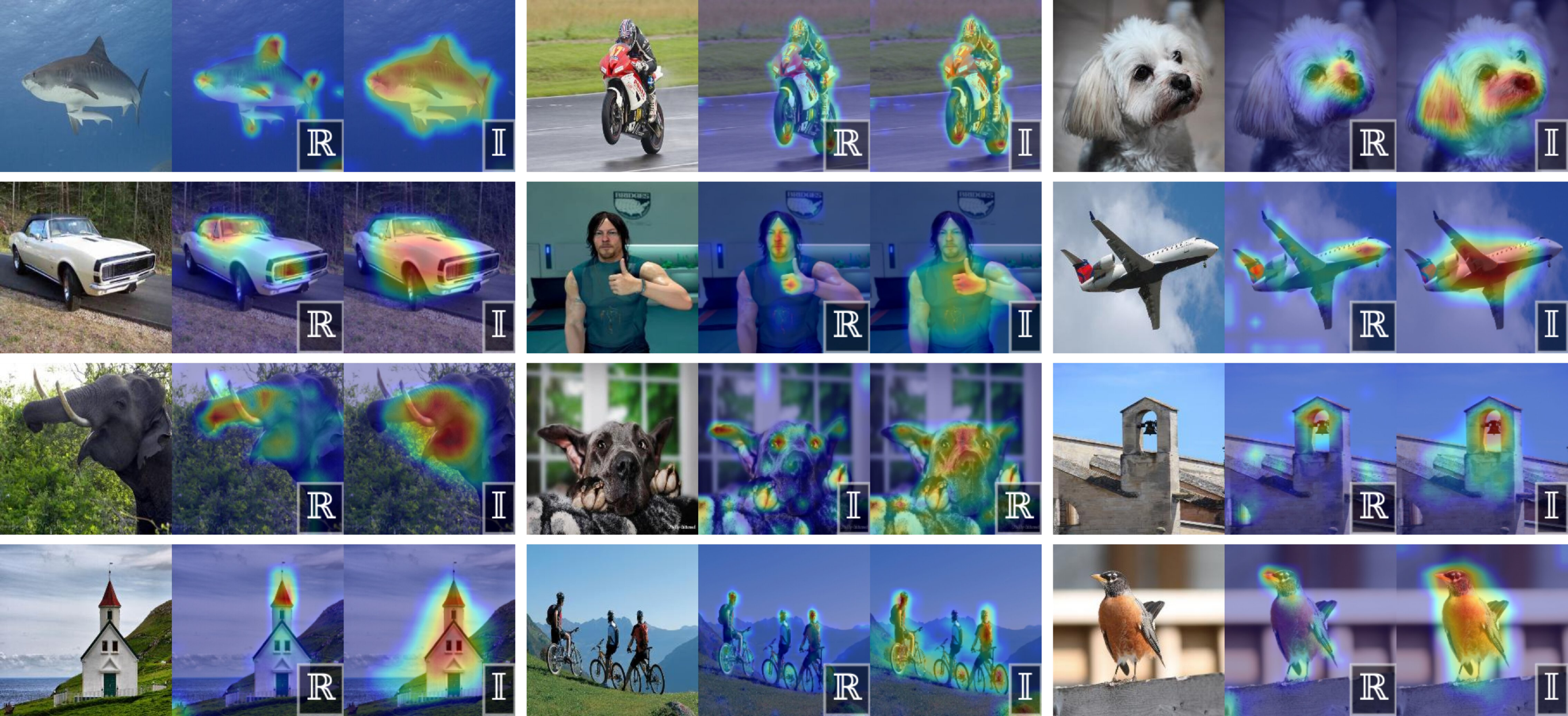}
    \caption{Assorted activation maps (via KPCA-CAM) for mid-level Phasor Blocks of PsychoDW. Real and imaginary components are denoted by $\mathbb{R}$ and $\mathbb{I}$.}
    \label{fig:psychodw_kpcacams}
\end{figure}

\begin{figure}
    \centering
    \includegraphics[width=1\linewidth]{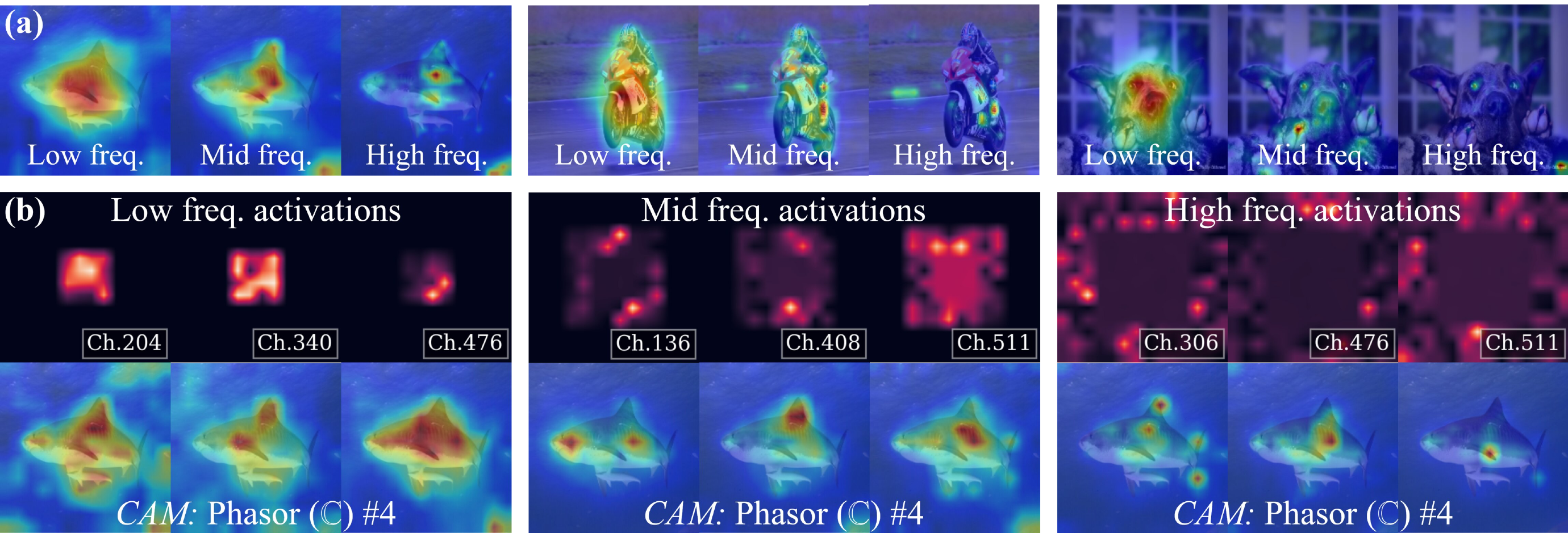}
    \caption{PsychoDW Phasor Block salience maps (via HiResCAM) conditioned on gradients \textbf{(a)} from individual Spectral Branch sub-bands and \textbf{(b)} from individual frequency domain feature channels.}
    \label{fig:psychodw_grad_cams}
\end{figure}

Figures \ref{fig:psychodw_vs_convnext} through \ref{fig:psychodw_grad_cams} present qualitative visualisations and analysis for PsychoDW identical to those applied to Psycho-B we presented in Section \ref{sec:experiments}. Overall, these show similar results: 

\begin{itemize}
    \item Figure \ref{fig:psychodw_vs_convnext} \textbf{(a)} shows that PsychoDW's \ac{DVC} filters also learn sparse selections of frequencies across each sub-band.
    \item Figure \ref{fig:psychodw_vs_convnext} \textbf{(b)} shows that similar to the Psycho-B vs. ResNet-101 comparison in Figure \ref{fig:psycho_vs_resnet_results} \textbf{(b)}, salience maps of PsychoDW's low-mid level Phasor Blocks clearly emphasis specific object parts, while those of ConvNeXt-S are much more general and diffuse. Further examples of the former are shown in Figure \ref{fig:psychodw_kpcacams}.
    \item Figure \ref{fig:psychodw_grad_cams} shows that similar to for Psycho-B in Figure \ref{fig:grad_cams}, PsychoDW's \ac{DVC} appears to distribute object parts by scale between the three sub-bands, and individual filters within each sub-band target distinct selections of object parts. 
\end{itemize}

Overall, these results are highly consistent with those for Psycho-B, showing that \ac{DVC} abstractions and object-part-centric Phasor Block representations also translate to ConvNeXt-S.

\newpage
\FloatBarrier

\FloatBarrier
\newpage

\subsection{Smaller classification datasets}\label{sec:app_small_classification}
\textbf{CIFAR-10} is a small scale dataset comprising 50000 natural images for training and 10000 images for testing across 10 classes, at a resolution of $32\times 32$ \citep{cifar}. For compatibility with this lower resolution (the ImageNet models have $224 \times 224$ input resolution), we reduce initial downsampling steps from our models. For ResNet and ResNet-based PsychoNet models, we removed the first maxpooling layer and set stride=1 for the first two ResBlocks that originally had stride=2. For ConvNeXt-S and PsychoDW, we replace the initial $4\times 4$ patch embedding layer with a standard $3\times 3$ Conv2D layer, and set stride=1 for the second downsampling layer. Table \ref{tab:cifar10_recipe} presents the training recipe for the CIFAR-10 experiments.

\begin{table}[ht]
    \caption{CIFAR-10 training recipe}
    \label{tab:cifar10_recipe}\textbf{}
    \centering
    \footnotesize
    \begin{tabularx}{0.85\textwidth}{l@{\hspace{3em}}X}
        \toprule
        Setting & Value \\
        \midrule

        Image size & $32\times 32$\\
        Epochs & 35 \\
        Batch size & 64 \\
        \midrule

        Loss & Cross entropy \\
        Optimizer & AdamW ($\beta_1=0.9, \beta_2=0.999$) \\
        Scheduler & OneCycle \\
        Learning rate (LR) & $10^{-3}$ \\
        \midrule

        Augmentation & crop, horizontal flip\\
        \midrule
        
        GPU & \makecell[l]{
            $1 \times$ NVIDIA A100: Psycho-S/B, Psycho-Eff-S/B, ResNet50/101\\
            $1 \times$ NVIDIA H100: All other models
         }\\
        \bottomrule
    \end{tabularx}
\end{table}

\textbf{CIFAR-100} contains the same images and train-test split as CIFAR-10, but with labels reorganised into 100 classes instead of 10. We use the same model configurations and training recipe as CIFAR-10, but increase the number of epochs to 90 since the greater number of classes results in a harder classification problem. Table \ref{tab:cifar10_recipe} presents the training recipe for the CIFAR-10 experiments. Overall, all of our PsychoNet models outperformed their respective \ac{CNN} baselines.

\textbf{ImageNet-100} is a subset of the ImageNet dataset \citep{imagenet} that contains examples for 100 classes. It contains 130100 images for training and 5100 images for testing, at the original resolution of $224 \times 224$. The model architectures remain the same as the ImageNet experiments, but with the output linear layer modified to predict 100 logits. We use the same training recipe as ImageNet-1K (Table \ref{tab:in1k_recipe}), but reduce the batch size to 128. Psycho-S/B, Psycho-Eff-S/B and ResNet50/101 were trained on $1\times$ NVIDIA A100, while all over models used $1\times$ AMD MI300X. Overall, the ResNet-based PsychoNet models outperformed their respective baselines, but PsychoDW fell slightly short of ConvNeXt-S. 

\subsection{Clustering visualisation}
\begin{figure}[ht]
    \centering
    \includegraphics[width=0.98\linewidth]{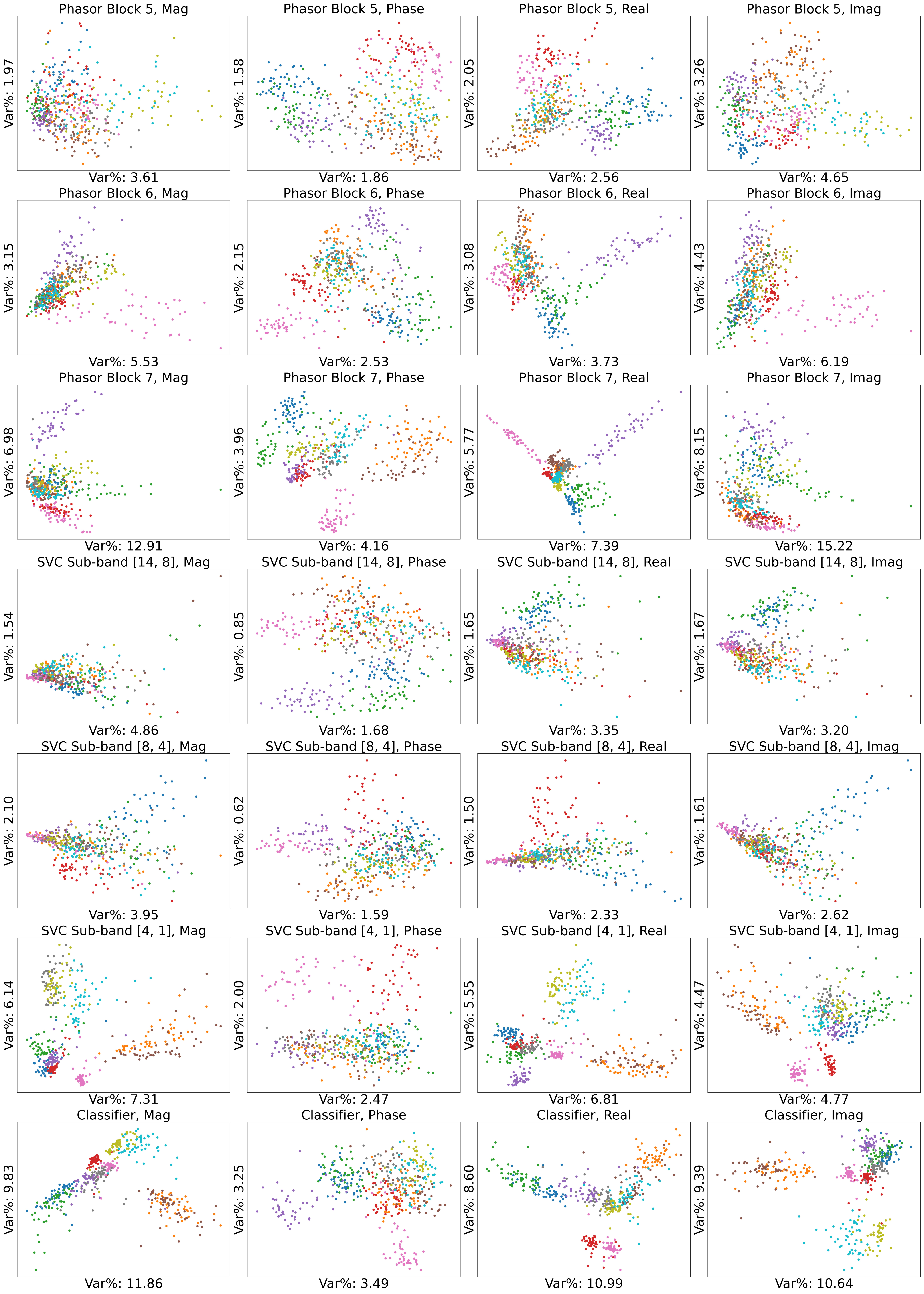}
    \caption{Psycho-B clustering visualisation.}
    \label{fig:clustering}
\end{figure}

Figure \ref{fig:clustering} presents an initial visualisation of clustering characteristics of Phasor Block activations and \ac{DVC} for Psycho-B. These show 2D PCA projections of features computed on samples from 10 randomly-selected classes from ImageNet-1K. Observable clustering emerges across both the real and imaginary/magnitude-phase feature components, and becomes increasingly pronounced at deeper layers.

\FloatBarrier

\section{Model configurations}\label{app:architecture}
Here we provide full details of the architectural configurations of all of our models. For all tables, we use the ResNet approach of counting the number of model layers as the number of convolutional and linear layers; each element-wise filter block in Hadamard Blocks are also counted as one layer.

\subsection{\ac{CNN} baselines}
The ResNet50, 101 and 152 models we use are from \citet{resnet} and are implemented in most common deep learning frameworks (we use the one from \texttt{PyTorch} \citep{pytorch}). For ResNet270, we follow the block configurations in \citet{revisiting_resnets}, but do not implement any of the newer blocks/layers they also introduce, so it purely just adds more residual bottleneck blocks (ResBlocks) to ResNet152 for fair scaling.  Table \ref{tab:resnet_configs} compares the sizes of the four ResNet models as well as their block configurations, grouped by feature resolution (which are $56\times 56, \ 28 \times 28, \ 14 \times 14$ and $7 \times 7$).

\begin{table}[ht]
    \footnotesize
    \caption{ResNet block configurations.}
    \label{tab:resnet_configs}
    \centering
    \begin{tabularx}{0.7\textwidth}{lXXX}
         \toprule
         Model & Parameters (M) & \# Layers & \# Blocks\\
         \midrule
         ResNet50 & 25.56 & 54 & \texttt{[3-4-6-3]} \\
         ResNet101 & 44.55 & 105 & \texttt{[3-4-23-3]} \\
         ResNet152 & 60.19 & 156 & \texttt{[3-8-36-3]} \\
         ResNet270 & 89.60 & 276 & \texttt{[4-29-53-3]} \\
         \bottomrule
    \end{tabularx}
\end{table}

For ConvNeXt-S, we follow the original implemention in \citet{convnext}.

\subsection{PsychoNet}\label{app:psychonet}
Table \ref{tab:psycho_variants} summarises the key configuration details of each base PsychoNet variant, namely the feature resolution and channel width at each Phasor Block and the number of filters used in the \ac{DVC} module. Further architectural details for each model are reported in Tables \ref{tab:arch_psychos} through \ref{tab:arch_psychodw}. For Phasor Blocks, we list each layer using the `\textbf{resolution}: layer configuration' format. The ResNet-based PsychoNet models use the same initial input embedding layer as ResNet ($7\times 7$ Conv2D and maxpooling) is used, while PsychoDW uses the same $4\times4$ patch embeddings as ConvNeXt-S. Interestingly, we found that using the initial layers of ResNet50, instead of ConvNeXt-S, in our ConvNeXt-S based PsychoDW actually yielded better results (approx. $\uparrow$ 0.5\% top-1 accuracy on ImageNet-1K), so we chose to use it for the model. However, we do change all ConvBlocks in \phasorc (see Figure \ref{fig:phasor_appendix}) to depthwise convolution blocks to maintain general faithfulness to the ConvNeXt model.

Finally, the companding operation we apply after taking the 2D \ac{FFT} (in Figure \ref{fig:architecture}) simply zeros the DC component and applies the element-wise function:
\begin{equation}
    x \in \mathbb{C}, \quad \text{Compand}: x \rightarrow |x|^{\frac{1}{1.25}} \cdot \exp(i \angle x)
\end{equation}
where $|x|$ denotes the magnitude of $x$ and $\angle x$ its phase. Since the exponent applied to the magnitude is $\in (0, 1)$, this function compresses frequencies of large magnitude (i.e. frequencies very close to the DC component), and expands the magnitude of those further from it.

\begin{table}[ht]
    \centering
    \caption{Configuration summary of different PsychoNet variants. For Phasor Blocks, we display $\textbf{resolution}$: [$\#$channels per block], and `$(\mathbb{I})$' denotes a Phasor Block $(\mathbb{I})$). $\#$Filters denotes the number of channels of element-wise filters per sub-band of \ac{DVC}. $\#$Layers show overall layers / complex convolution layer counts.}
    \label{tab:psycho_variants}
    \begin{tabularx}{1\textwidth}{lXrrr}
        \toprule
        Model & Phasor Blocks & $\#$Filters  & $\#$Layers &  Params (M)
        \\ \midrule
        
        Psycho-S
        & \makecell[tl]{
            $\mathbf{14 \times 14}$: $\sqr{\text{256 $(\mathbb{I})$, 256, 384, 512, 512}}$  
        }
        & 512
        & 65 / 9
        & 25.35
        \\ \midrule

        Psycho-B
        & \makecell[tl]{
            $\mathbf{28 \times 28}$: $\sqr{\text{256 $(\mathbb{I})$, 256, 256, 384}}$\\
            $\mathbf{14 \times 14}$: $\sqr{\text{384, 384, 512, 512, 512}}$\\ 
        }
        & 512
        & 93 / 13
        & 42.01
        \\ \midrule

        Psycho-L
        & \makecell[tl]{
            $\mathbf{28 \times 28}$: $\sqr{\text{256 $(\mathbb{I})$, 512, 512, 512}}$\\
            $\mathbf{14 \times 14}$: $\sqr{\text{512, 512, 512, 512, 512}}$\\ 
        }
        & 512
        & 93 / 13
        & 61.28
        \\ \midrule

        Psycho-H
        & \makecell[tl]{
            $\mathbf{28 \times 28}$: $\sqr{\text{256 $(\mathbb{I})$, 512, 512, 512}}$\\
            $\mathbf{14 \times 14}$: $\sqr{\text{512, 512, 512, 640, 1024}}$\\ 
        }
        & 1024
        & 93 / 13
        & 88.61
        \\ \midrule

        Psycho-DW
        & \makecell[tl] {
            $\mathbf{28 \times 28}$: $\sqr{\text{256 $(\mathbb{I})$, 256, 256, 512}}$\\
            $\mathbf{14 \times 14}$: $\sqr{\text{512, 1024, 1024, 1024, 1024}}$\\ 
        }
        & 2048
        & 109 / 13
        & 49.512
        \\ \bottomrule
    \end{tabularx}
\end{table}


\newpage
\begin{table}[ht]
    \footnotesize
    \centering
    \caption{Detailed architecture of Psycho-S.}
    \label{tab:arch_psychos}
    \begin{tabularx}{0.9\textwidth}{l@{\hspace{4em}}X}
        \toprule
        \multicolumn{2}{c}{\textbf{Psycho-S - comparable size to ResNet50}} \\
        \midrule

        Parameters (M) & 25.35 \\
        \# Layers (overall) & 65 \\
        \# Layers (complex) & 9 \\

        \midrule
        \multicolumn{2}{c}{\textbf{Blocks}} \\
        \midrule
         
        Input layer & Conv2D($7\times 7$, \din=3, \dout=64, stride=2), MaxPool($3\times3$, stride=2)\\
        \midrule

        Initial \ac{CNN} layers & First 7 ResBlocks from ResNet50  (first two resolution stages).
        \\
        \midrule

        Phasor Blocks 
        &
        \makecell[l]{
            $\bm{14\times14}$: $(\mathbb{I}) \sqr{\text{\din=128, \dout=256, stride=2}}$\\
            $\bm{14\times14}$: $(\mathbb{C}) \sqr{\text{\din=256, \dout=256}}$\\
            $\bm{14\times14}$: $(\mathbb{C}) \sqr{\text{\din=256, \dout=384}}$\\
            $\bm{14\times14}$: $(\mathbb{C}) \sqr{\text{\din=384, \dout=512}}$\\
            $\bm{14\times14}$: $(\mathbb{C}) \sqr{\text{\din=512, \dout=512}}$
        }
        \\
        
        \midrule

        Spectral filters &
        Sub-bands ([crop, drop]): [14, 8], [8, 4], [4, 1], \quad d\_filter = 512
        \\
        \midrule

        Output layer  &
        Average pool, ComplexLinear(\din=1536, \dout=1000), Softmax
        \\
        \bottomrule
    \end{tabularx}
\end{table}

\begin{table}[ht]
    \footnotesize
    \centering
    \caption{Detailed architecture of Psycho-B.}
    \label{tab:arch_psychob}
    \begin{tabularx}{0.9\textwidth}{l@{\hspace{4em}}X}
        \toprule
        \multicolumn{2}{c}{\textbf{Psycho-B architecture - comparable size to ResNet101}} \\
        \midrule

        Parameters (M) & 42.01 \\
        \# Layers (overall) & 93 \\
        \# Layers (complex) & 13 \\

        \midrule
        \multicolumn{2}{c}{\textbf{Blocks}} \\
        \midrule
         
        Input layer & Conv2D($7\times 7$, \din=3, \dout=64, stride=2), MaxPool($3\times3$, stride=2)\\
        \midrule

        Initial \ac{CNN} layers & First 7 ResBlocks from ResNet101 (first two resolution stages).
        \\
        \midrule

        Phasor Blocks 
        &
        \makecell[l]{
            $\bm{28\times28}$: $(\mathbb{I}) \sqr{\text{\din=128, \dout=256}}$\\
            $\bm{28\times28}$: $(\mathbb{C}) \sqr{\text{\din=256, \dout=256}}$\\
            $\bm{28\times28}$: $(\mathbb{C}) \sqr{\text{\din=256, \dout=256}}$\\
            $\bm{28\times28}$: $(\mathbb{C}) \sqr{\text{\din=256, \dout=384}}$\\
            \\
            $\bm{14\times14}$: $(\mathbb{C}) \sqr{\text{\din=384, \dout=384, stride=2}}$\\
            $\bm{14\times14}$: $(\mathbb{C}) \sqr{\text{\din=384, \dout=384}}$\\
            $\bm{14\times14}$: $(\mathbb{C}) \sqr{\text{\din=384, \dout=512}}$\\
            $\bm{14\times14}$: $(\mathbb{C}) \sqr{\text{\din=512, \dout=512}}$\\
            $\bm{14\times14}$: $(\mathbb{C}) \sqr{\text{\din=512, \dout=512}}$\\
        }
        \\
        
        \midrule

        Spectral filters &
        Sub-bands ([crop, drop]): [14, 8], [8, 4], [4, 1], \quad d\_filter = 512
        \\
        \midrule

        Output layer  &
        Average pool, ComplexLinear(\din=1536, \dout=1000), Softmax
        \\
        \bottomrule
    \end{tabularx}
\end{table}

\begin{table}[ht]
    \footnotesize
    \centering
    \caption{Detailed architecture of Psycho-L.}
    \label{tab:arch_psychol}
    \begin{tabularx}{0.9\textwidth}{l@{\hspace{4em}}X}
        \toprule
        \multicolumn{2}{c}{\textbf{Psycho-L architecture - comparable size to ResNet152}} \\
        \midrule

        Parameters (M) & 61.28 \\
        \# Layers (overall) & 93 \\
        \# Layers (complex) & 13 \\

        \midrule
        \multicolumn{2}{c}{\textbf{Blocks}} \\
        \midrule
         
        Input layer & Conv2D($7\times 7$, \din=3, \dout=64, stride=2), MaxPool($3\times3$, stride=2)\\
        \midrule

        Initial \ac{CNN} layers & First 7 ResBlocks from ResNet152.
        \\
        \midrule

        Phasor Blocks 
        &
        \makecell[l]{
            $\bm{28\times28}$: $(\mathbb{I}) \sqr{\text{\din=128, \dout=256}}$\\
            $\bm{28\times28}$: $(\mathbb{C}) \sqr{\text{\din=256, \dout=512}}$\\
            $\bm{28\times28}$: $(\mathbb{C}) \sqr{\text{\din=512, \dout=512}}$\\
            $\bm{28\times28}$: $(\mathbb{C}) \sqr{\text{\din=512, \dout=512}}$\\
            \\
            $\bm{14\times14}$: $(\mathbb{C}) \sqr{\text{\din=512, \dout=512, stride=2}}$\\
            $\bm{14\times14}$: $(\mathbb{C}) \sqr{\text{\din=512, \dout=512}}$\\
            $\bm{14\times14}$: $(\mathbb{C}) \sqr{\text{\din=512, \dout=512}}$\\
            $\bm{14\times14}$: $(\mathbb{C}) \sqr{\text{\din=512, \dout=512}}$\\
            $\bm{14\times14}$: $(\mathbb{C}) \sqr{\text{\din=512, \dout=512}}$\\
        }
        \\
        
        \midrule

        Spectral filters &
        Sub-bands ([crop, drop]): [14, 8], [8, 4], [4, 1], \quad d\_filter = 512
        \\
        \midrule

        Output layer  &
        Average pool, ComplexLinear(\din=1536, \dout=1000), Softmax
        \\
        \bottomrule
    \end{tabularx}
\end{table}

\begin{table}[ht]
    \footnotesize
    \centering
    \caption{Detailed architecture of Psycho-H.}
    \label{tab:arch_psychoh}
    \begin{tabularx}{0.9\textwidth}{l@{\hspace{4em}}X}
        \toprule
        \multicolumn{2}{c}{\textbf{Psycho-H architecture - comparable size to ResNet270}} \\
        \midrule

        Parameters (M) & 88.61 \\
        \# Layers (overall) & 93 \\
        \# Layers (complex) & 13 \\

        \midrule
        \multicolumn{2}{c}{\textbf{Blocks}} \\
        \midrule
         
        Input layer & Conv2D($7\times 7$, \din=3, \dout=64, stride=2), MaxPool($3\times3$, stride=2)\\
        \midrule

        Initial \ac{CNN} layers & First 7 ResBlocks from ResNet270.
        \\
        \midrule

        Phasor Blocks 
        &
        \makecell[l]{
            $\bm{28\times28}$: $(\mathbb{I}) \sqr{\text{\din=128, \dout=256}}$\\
            $\bm{28\times28}$: $(\mathbb{C}) \sqr{\text{\din=256, \dout=512}}$\\
            $\bm{28\times28}$: $(\mathbb{C}) \sqr{\text{\din=512, \dout=512}}$\\
            $\bm{28\times28}$: $(\mathbb{C}) \sqr{\text{\din=512, \dout=512}}$\\
            \\
            $\bm{14\times14}$: $(\mathbb{C}) \sqr{\text{\din=512, \dout=512, stride=2}}$\\
            $\bm{14\times14}$: $(\mathbb{C}) \sqr{\text{\din=512, \dout=512}}$\\
            $\bm{14\times14}$: $(\mathbb{C}) \sqr{\text{\din=512, \dout=512}}$\\
            $\bm{14\times14}$: $(\mathbb{C}) \sqr{\text{\din=512, \dout=640}}$\\
            $\bm{14\times14}$: $(\mathbb{C}) \sqr{\text{\din=640, \dout=1024}}$\\
        }
        \\
        
        \midrule

        Spectral filters &
        Sub-bands ([crop, drop]): [14, 8], [8, 4], [4, 1], \quad d\_filter = 1024
        \\
        \midrule

        Output layer  &
        Average pool, ComplexLinear(\din=3072, \dout=1000), Softmax
        \\
        \bottomrule
    \end{tabularx}
\end{table}

\begin{table}[ht]
    \footnotesize
    \centering
    \caption{Detailed architecture of PsychoDW.}
    \label{tab:arch_psychodw}
    \begin{tabularx}{0.9\textwidth}{l@{\hspace{4em}}X}
        \toprule
        \multicolumn{2}{c}{\textbf{PsychoDW architecture - comparable size to ConvNeXt-S}} \\
        \midrule

        Parameters (M) & 49.512 \\
        \# Layers (overall) & 109 \\
        \# Layers (complex) & 13 \\

        \midrule
        \multicolumn{2}{c}{\textbf{Blocks}} \\
        \midrule
         
        Input layer & Conv2D($7\times 7$, \din=3, \dout=64, stride=2), MaxPool($3\times3$, stride=2)\\
        \midrule

        Initial \ac{CNN} layers & First 7 ResBlocks from ResNet50.
        \\
        \midrule

        Phasor Blocks 
        &
        \makecell[l]{
            $\bm{28\times28}$: $(\mathbb{I}) \sqr{\text{\din=128, \dout=256}}$\\
            $\bm{28\times28}$: $(\mathbb{C}) \sqr{\text{\din=256, \dout=256}}$\\
            $\bm{28\times28}$: $(\mathbb{C}) \sqr{\text{\din=256, \dout=256}}$\\
            $\bm{28\times28}$: $(\mathbb{C}) \sqr{\text{\din=256, \dout=512}}$\\
            \\
            $\bm{14\times14}$: $(\mathbb{C}) \sqr{\text{\din=512, \dout=512, stride=2}}$\\
            $\bm{14\times14}$: $(\mathbb{C}) \sqr{\text{\din=512, \dout=1024}}$\\
            $\bm{14\times14}$: $(\mathbb{C}) \sqr{\text{\din=1024, \dout=1024}}$\\
            $\bm{14\times14}$: $(\mathbb{C}) \sqr{\text{\din=1024, \dout=1024}}$\\
            $\bm{14\times14}$: $(\mathbb{C}) \sqr{\text{\din=1024, \dout=1024}}$\\
        }
        \\
        
        \midrule

        Spectral filters &
        Sub-bands ([crop, drop]): [14, 8], [8, 4], [4, 1], \quad d\_filter = 1024
        \\
        \midrule

        Output layer  &
        Average pool, ComplexLinear(\din=3072, \dout=1000), Softmax
        \\
        \bottomrule
    \end{tabularx}
\end{table}
\FloatBarrier

\subsection{Phasor Block architecture}\label{app:phasor_block}
\begin{figure}[ht]
    \centering
    \includegraphics[width=0.98\linewidth]{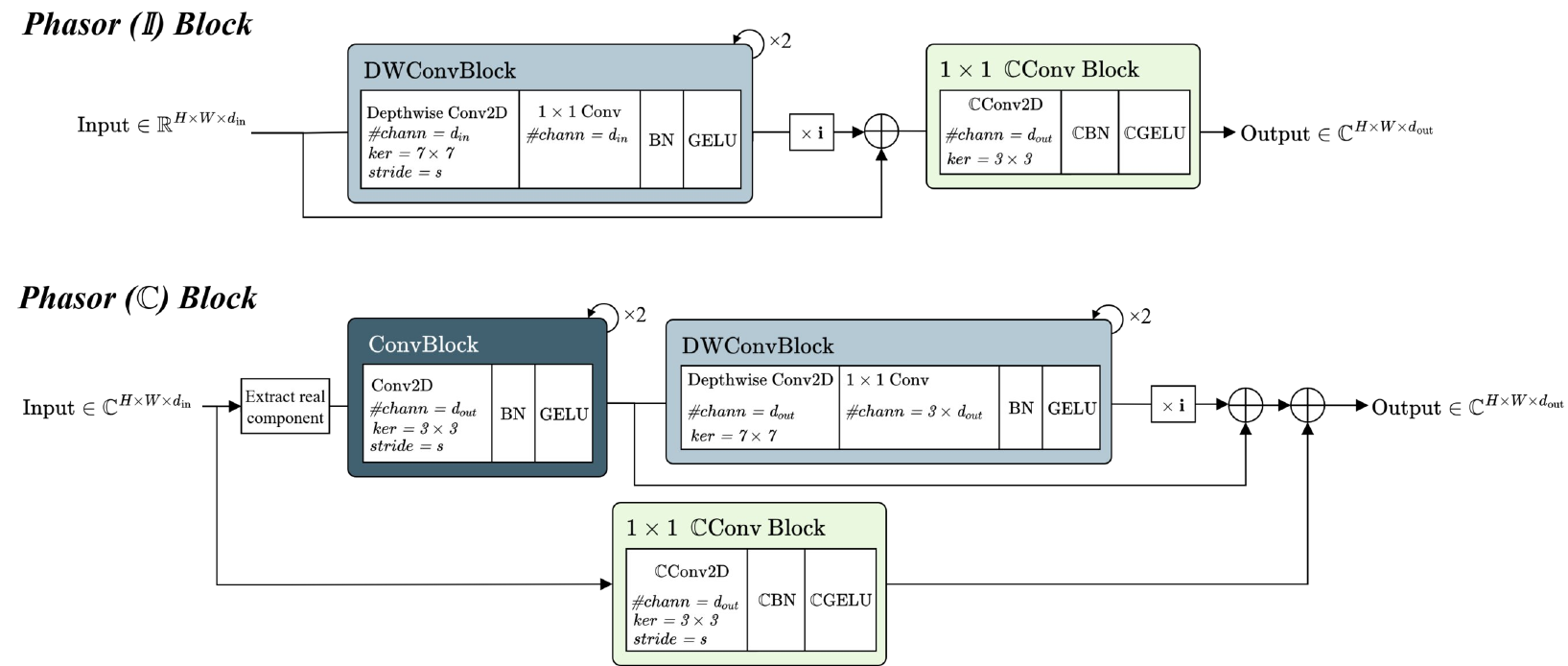}
    \caption{Further architecture details for the Phasor Blocks presented in Figure \ref{fig:architecture}. For ConvNeXt-based PsychoNet, we replace the two ConvBlocks at the start of \phasorc blocks with two DWConvBlocks with the same number of channels. $\mathbb{C}$Conv/BN/GELU denote complex-valued convolution, batch norm and GELU operations - see Appendix \ref{app:background_complex}. The following PsychoNet architecture tables specify the values of \din, \dout and stride ($s$) for all of their Phasor Blocks.}
    \label{fig:phasor_appendix}
\end{figure}

Figure \ref{fig:phasor_appendix} provides detailed architectural diagrams of Phasor Blocks, with key design choices discussed below.

\phasori blocks generate an initial set of imaginary components using depthwise convolution (`DWConv') blocks, comprising pairs of depthwise and $1\times 1$ convolution layers. This configuration decouples spatial and channel mixing, which is intended to encourage cross-channel interactions without interfering with spatial relationships. In natural complex signals, the real and imaginary components carry complementary information for the same spatial location \citep{imageprocessingbook,complex_survey}, so it is likely important that our generated imaginary features do not significantly introduce new spatial information. A $1 \times 1$ complex convolution block then mixes the real and imaginary features. Subsequently, \phasorc blocks further refine the complex representations. The top branch generates new real and imaginary features, while the bottom channel-mixes the original features and combines them with the new ones. For ConvNeXt-based PsychoNet, we replace \phasorc's regular convolution (`Conv') blocks with further DWConv blocks with $7\times7$ kernel size, matching ConvNeXt's main computational block.

\subsection{Efficient PsychoNet}
The base PsychoNet models have considerably higher FLOP count compared to ResNet primarily because they only downsample spatial resolution to $14\times 14$ at smallest, whereas ResNet (and many other standard \acp{CNN}) typically downsample further to $7\times 7$. To address this, we introduced the FLOP-efficient PsychoNet variants Psycho-Eff-S/B/L, which incorporate two main architectural changes to improve computational efficiency. First, we add an additional downsampling stage to the Phasor Blocks to reduce the feature resolution to $7\times 7$. In these models, the $[14, 7]$ sub-band is extracted from the last $14\times 14$ Phasor Block, while the remaining sub-bands are extracted from the final $7\times 7$ block. The specific Phasor Blocks from which these sub-bands are extracted for \ac{DVC} are explicitly labelled in the architecture tables below. Second, within the Phasor $(\mathbb{C})$ blocks, we streamline the architecture by using only one Conv Block and one DWConv block respectively, whereas the base PsychoNet models utilize two of each (as shown in Figure \ref{fig:architecture}). Table \ref{tab:psycho_eff_variants} summarises the key configuration details of each efficient PsychoNet variant, namely the feature resolution and channel width at each Phasor Block and the number of filters used in the \ac{DVC} module. Further architectural details for each model are reported in Tables \ref{tab:arch_psychoeffs} through \ref{tab:arch_psychoeffl}.

\begin{table}[ht]
    \centering
    \caption{Configuration summary of different efficient PsychoNet variants. For Phasor Blocks, we display $\textbf{resolution}$: [$\#$channels per block], and `$(\mathbb{I})$' denotes a Phasor Block $(\mathbb{I})$). $\#$Filters denotes the number of channels of element-wise filters per sub-band of \ac{DVC}. $\#$Layers show overall layers / complex convolution layer counts.}
    \label{tab:psycho_eff_variants}
    \begin{tabularx}{1\textwidth}{lXrrr}
        \toprule
        Model & Phasor Blocks & $\#$Filters  & $\#$Layers &  Params (M)
        \\ \midrule
        
        Psycho-Eff-S
        & \makecell[tl]{
            $\mathbf{14 \times 14}$: $\sqr{\text{256 $(\mathbb{I})$, 256, 512}}$\\
            \hspace{0.855em} $\mathbf{7 \times 7}$: $\sqr{512, 512, 768}$  
        }
        & 768
        & 57 / 9
        & 25.35
        \\ \midrule

        Psycho-Eff-B
        & \makecell[tl]{
            $\mathbf{28 \times 28}$: $\sqr{\text{256 $(\mathbb{I})$}}$\\
            $\mathbf{14 \times 14}$: $\sqr{512, 512, 512, 512}$\\
            \hspace{0.855em} $\mathbf{7 \times 7}$: $\sqr{768, 768, 768}$
        }
        & 768
        & 65 / 11
        & 45.82
        \\ \midrule

        Psycho-Eff-L
        & \makecell[tl]{
            $\mathbf{28 \times 28}$: $\sqr{\text{256 $(\mathbb{I})$}}$\\
            $\mathbf{14 \times 14}$: $\sqr{512, 512, 512, 512, 512, 512, 512}$\\
            \hspace{0.855em} $\mathbf{7 \times 7}$: $\sqr{1024, 1024}$
        }
        & 1024
        & 85 / 13
        & 62.03
        \\ \midrule
    \end{tabularx}
\end{table}

\begin{table}[ht]
    \footnotesize
    \centering
    \caption{Detailed architecture of Psycho-Eff-S.}
    \label{tab:arch_psychoeffs}
    \begin{tabularx}{0.9\textwidth}{l@{\hspace{4em}}X}
        \toprule
        \multicolumn{2}{c}{\textbf{Psycho-Eff-S}} \\
        \midrule

        Parameters (M) & 25.35 \\
        \# Layers (overall) & 57 \\
        \# Layers (complex) & 9 \\

        \midrule
        \multicolumn{2}{c}{\textbf{Blocks}} \\
        \midrule
         
        Input layer & Conv2D($7\times 7$, \din=3, \dout=64, stride=2), MaxPool($3\times3$, stride=2)\\
        \midrule

        Initial \ac{CNN} layers & First 7 ResBlocks from ResNet50.
        \\
        \midrule

        Phasor Blocks 
        &
        \makecell[l]{
            $\bm{14\times14}$: $(\mathbb{I}) \sqr{\text{\din=128, \dout=256, stride=2}}$\\
            $\bm{14\times14}$: $(\mathbb{C}) \sqr{\text{\din=256, \dout=256}}$\\
            $\bm{14\times14}$: $(\mathbb{C}) \sqr{\text{\din=256, \dout=512}}$ \quad $\rightarrow$ extract [14, 7] sub-band\\
            \\
            $\bm{7\times7}$: \hspace{0.5em} $(\mathbb{C}) \sqr{\text{\din=512, \dout=512, stride=2}}$\\
            $\bm{7\times7}$: \hspace{0.5em} $(\mathbb{C}) \sqr{\text{\din=512, \dout=512}}$\\
            $\bm{7\times7}$: \hspace{0.5em} $(\mathbb{C}) \sqr{\text{\din=512, \dout=768}}$ \quad $\rightarrow$ extract [7, 4], [4, 1] sub-bands
        }
        \\
        
        \midrule

        Spectral filters &
        Sub-bands ([crop, drop]): [14, 7], [7, 4], [4, 1]
        \\
        \midrule

        Output layer  &
        Average pool, ComplexLinear(\din=2304, \dout=1000), Softmax
        \\
        \bottomrule
    \end{tabularx}
\end{table}

\begin{table}[ht]
    \footnotesize
    \centering
    \caption{Detailed architecture of Psycho-Eff-B.}
    \label{tab:arch_psychoeffb}
    \begin{tabularx}{0.9\textwidth}{l@{\hspace{4em}}X}
        \toprule
        \multicolumn{2}{c}{\textbf{Psycho-Eff-B - efficient variant of Psycho-B}} \\
        \midrule

        Parameters (M) & 45.82 \\
        \# Layers (overall) & 65 \\
        \# Layers (complex) & 11 \\

        \midrule
        \multicolumn{2}{c}{\textbf{Blocks}} \\
        \midrule
         
        Input layer & Conv2D($7\times 7$, \din=3, \dout=64, stride=2), MaxPool($3\times3$, stride=2)\\
        \midrule

        Initial \ac{CNN} layers & First 7 ResBlocks from ResNet101.
        \\
        \midrule

        Phasor Blocks 
        &
        \makecell[l]{
            $\bm{28\times28}$: $(\mathbb{I}) \sqr{\text{\din=128, \dout=256}}$\\
            \\
            $\bm{14\times14}$: $(\mathbb{C}) \sqr{\text{\din=256, \dout=512, stride=2}}$\\
            $\bm{14\times14}$: $(\mathbb{C}) \sqr{\text{\din=512, \dout=512}}$\\
            $\bm{14\times14}$: $(\mathbb{C}) \sqr{\text{\din=512, \dout=512}}$\\
            $\bm{14\times14}$: $(\mathbb{C}) \sqr{\text{\din=512, \dout=512}}$ \quad $\rightarrow$ extract [14, 7] sub-band\\
            \\
            $\bm{7\times7}$: \hspace{0.5em} $(\mathbb{C}) \sqr{\text{\din=512, \dout=768, stride=2}}$\\
            $\bm{7\times7}$: \hspace{0.5em} $(\mathbb{C}) \sqr{\text{\din=768, \dout=768}}$\\
            $\bm{7\times7}$: \hspace{0.5em} $(\mathbb{C}) \sqr{\text{\din=768, \dout=768}}$ \quad $\rightarrow$ extract [7, 4], [4, 1] sub-bands
        }
        \\
        
        \midrule

        Spectral filters &
        Sub-bands ([crop, drop]): [14, 7], [7, 4], [4, 1]
        \\
        \midrule

        Output layer  &
        Average pool, ComplexLinear(\din=2304, \dout=1000), Softmax
        \\
        \bottomrule
    \end{tabularx}
\end{table}

\begin{table}[ht]
    \footnotesize
    \centering
    \caption{Detailed architecture of Psycho-Eff-L.}
    \label{tab:arch_psychoeffl}
    \begin{tabularx}{0.9\textwidth}{l@{\hspace{4em}}X}
        \toprule
        \multicolumn{2}{c}{\textbf{Psycho-Eff-L - efficient variant of Psycho-L}} \\
        \midrule

        Parameters (M) & 62.03 \\
        \# Layers (overall) & 85 \\
        \# Layers (complex) & 13 \\

        \midrule
        \multicolumn{2}{c}{\textbf{Blocks}} \\
        \midrule
         
        Input layer & Conv2D($7\times 7$, \din=3, \dout=64, stride=2), MaxPool($3\times3$, stride=2)\\
        \midrule

        Initial \ac{CNN} layers & First 11 ResBlocks from ResNet152.
        \\
        \midrule

        Phasor Blocks 
        &
        \makecell[l]{
            $\bm{28\times28}$: $(\mathbb{I}) \sqr{\text{\din=128, \dout=256}}$\\
            \\
            $\bm{14\times14}$: $(\mathbb{C}) \sqr{\text{\din=256, \dout=512, stride=2}}$\\
            $\bm{14\times14}$: $(\mathbb{C}) \sqr{\text{\din=512, \dout=512}}$\\
            $\bm{14\times14}$: $(\mathbb{C}) \sqr{\text{\din=512, \dout=512}}$\\
            $\bm{14\times14}$: $(\mathbb{C}) \sqr{\text{\din=512, \dout=512}}$\\
            $\bm{14\times14}$: $(\mathbb{C}) \sqr{\text{\din=512, \dout=512}}$\\
            $\bm{14\times14}$: $(\mathbb{C}) \sqr{\text{\din=512, \dout=512}}$\\
            $\bm{14\times14}$: $(\mathbb{C}) \sqr{\text{\din=512, \dout=512}}$ \quad $\rightarrow$ extract [14, 7] sub-band\\
            \\
            $\bm{7\times7}$: \hspace{0.5em} $(\mathbb{C}) \sqr{\text{\din=512, \dout=1024, stride=2}}$\\
            $\bm{7\times7}$: \hspace{0.5em} $(\mathbb{C}) \sqr{\text{\din=1024, \dout=1024}}$ \quad $\rightarrow$ extract [7, 4], [4, 1] sub-bands
        }
        \\
        
        \midrule

        Spectral filters &
        Sub-bands ([crop, drop]): [14, 7], [7, 4], [4, 1]
        \\
        \midrule

        Output layer  &
        Average pool, ComplexLinear(\din=3072, \dout=1000), Softmax
        \\
        \bottomrule
    \end{tabularx}
\end{table}

\newpage
\section{Ablation studies}\label{app:ablation}
\paragraph{CIFAR-10}
Below we present architecture details for the barebones ablation architectures used for the experiments in Table \ref{tab:barebones_ablation}. Table \ref{tab:arch_ablation_i} presents the architecture of Model I; Model II just removes the \ac{DVC} entirely, directly feeding the features from the Phasor Blocks into the output layer after applying \ac{FFT}. Table \ref{tab:arch_ablation_iii} presents the architecture of Model III, which replaces the Phasor Blocks of Model I with simple Conv. blocks, matching the overall parameter size and layer depth of Model I. Each simple Conv. block contains two stacks of $3\times3 \textrm{ Conv2D} \rightarrow \textrm{BatchNorm2D} \rightarrow \textrm{GeLU}$. Model IV removes \ac{DVC} from Model III in the same manner as Model I $\rightarrow$ Model II.

\begin{table}[ht]
    \footnotesize
    \centering
    \caption{Detailed architecture of Ablation Model I.}
    \label{tab:arch_ablation_i}
    \begin{tabularx}{0.9\textwidth}{l@{\hspace{4em}}X}
        \toprule
        \multicolumn{2}{c}{\textbf{Ablation Model I}} \\
        \midrule
        Parameters (M) & 2.366 \\
        \# Layers (overall) & 16 \\
        \# Layers (complex) & 5 \\
        \midrule
        \multicolumn{2}{c}{\textbf{Blocks}} \\
        \midrule
         
        Input layer & Conv2D($3\times 3$, \din=3, \dout=64, stride=1), No MaxPool\\
        \midrule
        Phasor Blocks 
        &
        \makecell[l]{
            $(\mathbb{I}) \sqr{\text{\din=64, \dout=256, stride=2}}$\\
            $(\mathbb{C}) \sqr{\text{\din=256, \dout=256}}$\\
        }
        \\
        
        \midrule
        Spectral filters &
        Sub-bands ([crop, drop]): [8, 4], [4, 1], \quad d\_filter = 256
        \\
        \midrule
        Output layer  &
        Average pool, ComplexLinear(\din=512, \dout=10), Softmax
        \\
        \bottomrule
    \end{tabularx}
\end{table}
\begin{table}[ht]
    \footnotesize
    \centering
    \caption{Detailed architecture of Ablation Model III.}
    \label{tab:arch_ablation_iii}
    \begin{tabularx}{0.9\textwidth}{l@{\hspace{4em}}X}
        \toprule
        \multicolumn{2}{c}{\textbf{Ablation Model III}} \\
        \midrule
        Parameters (M) & 2.360 \\
        \# Layers (overall) & 17 \\
        \# Layers (complex) & 1 \\
        \midrule
        \multicolumn{2}{c}{\textbf{Blocks}} \\
        \midrule
         
        Input layer & Conv2D($3\times 3$, \din=3, \dout=64, stride=1), No MaxPool\\
        \midrule
        Simple Conv. blocks & 
        \makecell[l]{
            $\sqr{\text{\din=64, \dout=64}}$\\
            $\sqr{\text{\din=64, \dout=64}}$\\
            $\sqr{\text{\din=64, \dout=128, stride=2}}$\\
            $\sqr{\text{\din=128, \dout=192}}$\\
            $\sqr{\text{\din=192, \dout=256}}$\\
        }\\
        \midrule
        Phasor Blocks & None \\
        \midrule
        Spectral filters &
        Sub-bands ([crop, drop]): [8, 4], [4, 1], \quad d\_filter = 256
        \\
        \midrule
        Output layer  &
        Average pool, ComplexLinear(\din=512, \dout=10), Softmax
        \\
        \bottomrule
    \end{tabularx}
\end{table}

\paragraph{ImageNet-1K ablations}
Below we present architecture details for the ablation experiments presented in Table \ref{tab:imagenet_ablation}. 

Model A is identical to Psycho-L (Table \ref{tab:arch_psychol}). Model B just removes Spectral Branches from Model A, replacing it with a single Hadamard Block with full band $14\times14$ filters and no prior DropCrop operations. Model C removes all Phasor $(\mathbb{C})$ Blocks and makes up for the resultant layer and parameter deficit by adding additional ResBlocks; its architecture is presented in Table \ref{tab:arch_ablation_c}. Model D removes Spectral Branches from Model C in the same manner as Model A $\rightarrow$ Model B.

\begin{table}[t]
    \footnotesize
    \centering
    \caption{Detailed architecture of the Model C ablation model.}
    \label{tab:arch_ablation_c}
    \begin{tabularx}{0.9\textwidth}{l@{\hspace{4em}}X}
        \toprule
        \multicolumn{2}{c}{\textbf{Ablation Model C architecture}} \\
        \midrule

        Parameters (M) & 60.42 \\
        \# Layers (overall) & 90 \\
        \# Layers (complex) & 2 \\

        \midrule
        \multicolumn{2}{c}{\textbf{Blocks}} \\
        \midrule
         
        Input layer & Conv2D($7\times 7$, \din=3, \dout=64, stride=2), MaxPool($3\times3$, stride=2)\\
        \midrule

        ResBlocks
        &
        \makecell[l]{
            $\bm{56\times56}$: $\sqr{\text{\din=64, \dbot=256, \dout=256}}$\\
            $\bm{56\times56}$: $\sqr{\text{\din=256, \dbot=64, \dout=256}} \times 2$\\
            $\bm{28\times28}$: $\sqr{\text{\din=256, \dbot=128, \dout=512, stride=2}}$\\
            $\bm{28\times28}$: $\sqr{\text{\din=512, \dbot=128, \dout=512}} \times 5$\\
            $\bm{28\times28}$: $\sqr{\text{\din=512, \dbot=256, \dout=1024}}$\\
            $\bm{28\times28}$: $\sqr{\text{\din=1024, \dbot=256, \dout=1024}} \times 6$\\
            $\bm{14\times14}$: $\sqr{\text{\din=1024, \dbot=512, \dout=2048, stride=2}}$\\
            $\bm{14\times14}$: $\sqr{\text{\din=2048, \dbot=512, \dout=2048}} \times 7$\\
            $\bm{14\times14}$: $\sqr{\text{\din=2048, \dbot=128, \dout=512}}$\\
        }
        \\
        \midrule

        Phasor Blocks 
        &
        \makecell[l]{
            $\bm{14\times14}$: $(\mathbb{I}) \sqr{\text{\din=512, \dout=512, stride=1}}$\\
        }
        \\
        
        \midrule

        \ac{DVC} filters &
        Sub-bands ([crop, drop]): [14, 8], [8, 4], [4, 1], \quad $d_\text{filter}$ 512
        \\
        \midrule

        Output layer  &
        Average pool, ComplexLinear(\din=1536, \dout=1000), Softmax
        \\
        \bottomrule
    \end{tabularx}
\end{table}

\end{document}